%% file: main.tex
\documentclass[10pt,twocolumn,letterpaper]{article}

\usepackage{cvpr}              

\input{preamble}

\definecolor{cvprblue}{rgb}{0.21,0.49,0.74}
\usepackage[pagebackref,breaklinks,colorlinks,allcolors=cvprblue]{hyperref}
\usepackage{capt-of} 
\usepackage[accsupp]{axessibility}  
\usepackage{xspace}
\newcommand{\sysname}{\texttt{M4Human}\xspace}

\def\paperID{***} 
\def\confName{CVPR}
\def\confYear{2026}

\title{M4Human: A Large-Scale Multimodal mmWave Radar Benchmark for \\  Human Mesh Reconstruction}

\author{
Junqiao Fan$^{1}$ \quad
Yunjiao Zhou$^{1}$ \quad
Yizhuo Yang$^{1}$ \quad
Xinyuan Cui$^{2,3}$ \quad
Jiarui Zhang$^{1}$ \\
Lihua Xie$^{1}$ \quad
Jianfei Yang$^{1}$ \quad
Chris Xiaoxuan Lu$^{4}$ \quad
Fangqiang Ding$^{2,5}$\thanks{Corresponding author. Email: \texttt{fding@mit.edu}}  \\
$^{1}$NTU \quad
$^{2}$University of Edinburgh \quad
$^{3}$UPenn \quad
$^{4}$UCL \quad 
$^{5}$MIT 
\vspace{-3em}
}

\begin{document}
\maketitle

\begin{strip}
  \centering
  \vspace{-2em}
  \includegraphics[width=1.\linewidth]{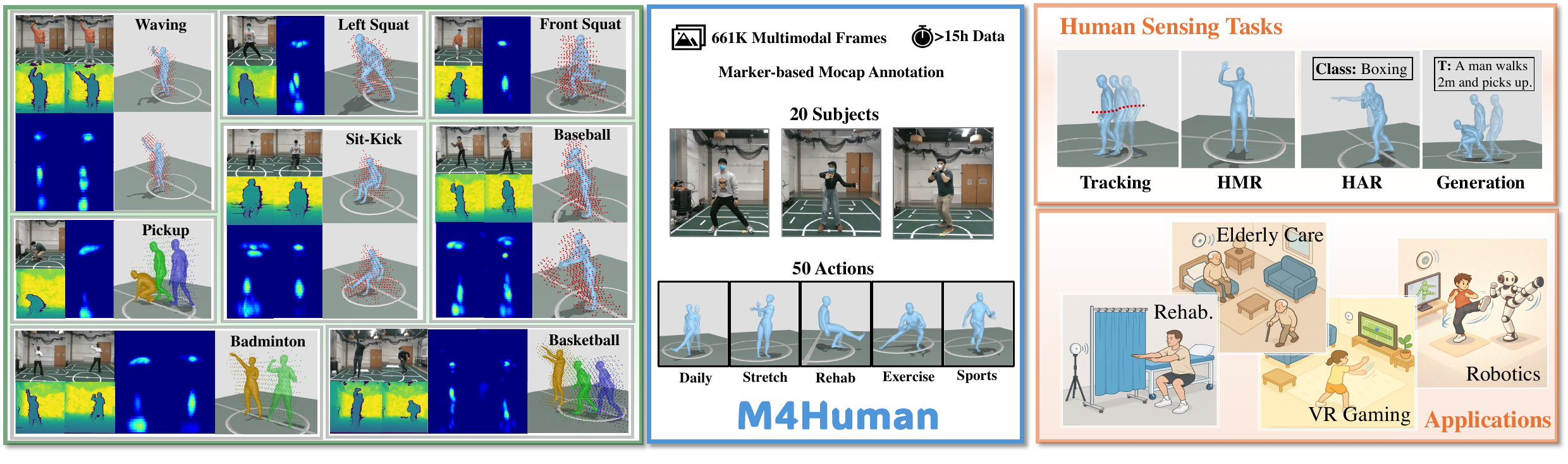}
  \captionof{figure}{Overview of \sysname, the \textbf{largest} multimodal dataset for high-fidelity mmWave radar-based human motion sensing. It covers diverse free-space motions (\eg, rehabilitation, exercise, and sports) beyond simple in-place actions, with high-quality marker-based motion annotations. Such diversity supports a broad range of human sensing tasks, including tracking, human mesh recovery, action recognition, and human motion generation, as well as privacy-preserving applications in elderly care, rehabilitation, robotics, and VR gaming.}
  
\end{strip}

\input{sec/0_abstract}    
\input{sec/1_intro}

\input{sec/2_related_work}

\input{sec/3_method}

\input{sec/4_experiment}

\input{sec/4_exp_tables}
\input{sec/5_discussion}
\input{sec/6_conclusion}

{
    \small
    \bibliographystyle{ieeenat_fullname}
    \bibliography{main}
}

\input{sec/7_supp}

\end{document}

%% file: preamble.tex
\usepackage{booktabs}
\usepackage{multicol}
\usepackage{multirow}
\usepackage{colortbl}
\usepackage{xcolor}
\usepackage{caption}
\usepackage{tabularray}
\usepackage{rotating}
\usepackage{makecell}
\usepackage{tabularx}
\usepackage{float}
\usepackage{adjustbox}
\usepackage{anyfontsize}
\usepackage{mathtools}
\usepackage{amsmath}
\usepackage{indentfirst}
\usepackage{float}

\usepackage{amssymb}
\usepackage{pifont}
\newcommand{\cmark}{\ding{51}}%
\usepackage{microtype}
\usepackage{cuted}     

%% file: sec/0_abstract.tex
\begin{abstract}


Human mesh reconstruction (HMR) provides direct insights into body-environment interaction, enabling various immersive applications. However, existing large-scale HMR benchmarks largely rely on line-of-sight RGB sensing, causing HMR systems to inherit the limitations of vision-based systems, including sensitivity to occlusion, lighting variation, and privacy concerns. These limitations have motivated growing interest in radio-frequency (RF) mmWave radar as a privacy-preserving and robust modality for human sensing. Despite this promise, current radar datasets remain limited by sparse skeleton annotations, small scale, and simple in-place actions. To address this gap, we introduce M4Human, the largest-scale multimodal benchmark to date for radar-based HMR, featuring 661K frames---$9\times$ larger than the previous largest---with high-resolution mmWave radar, RGB, and depth data. M4Human provides both raw radar tensors (RT) and processed radar point clouds (RPC), enabling research across different levels of RF signal granularity. It also includes high-quality motion capture (MoCap) annotations with 3D meshes and global trajectories, covering 20 subjects and 50 diverse actions, including in-place, seated, and free-space sports or rehabilitation movements. We establish benchmarks on RT and RPC modalities, as well as multimodal fusion with RGB-D inputs. Extensive results demonstrate the value of M4Human for radar-based human modeling while revealing persistent challenges under fast and unconstrained motion. Our benchmark is publicly available at {\href{https://fanjunqiao.github.io/M4Human-site/}{https://fanjunqiao.github.io/M4Human-site/}}.



\end{abstract}

%% file: sec/1_intro.tex
\vspace{-1em}
\section{Introduction}

Perceiving 3D human motion is essential for human-centric Physical AI systems to understand users' behavior and respond to their needs. A key enabler is markerless 3D human pose and shape regression, a core human sensing task that supports many applications such as human-robot interaction, rehabilitation, healthcare monitoring, surveillance, and VR/AR~\cite{stenum2021applications,avogaro2023markerless}. Classic human pose estimation (HPE) represents the human body as a graph of keypoints and infers their 3D locations~\cite{zheng2023deep}.  Moving beyond sparse landmarks, human mesh reconstruction (HMR) estimates both pose and shape of the human body as a dense surface mesh that can be used to model human-environment contact, enabling classical applications such as VR fitness gaming, virtual try-on, avatar creation, and computer-assisted coaching~\cite{liu2024deep,tian2023recovering}. The rapid rise of embodied AI and robotics~\cite{harlow2024new,gu2025humanoid} further make accurate HMR indispensable for safer human-robot interaction~\cite{drolet2023learning}, more reliable motion retargeting~\cite{ye2024skinned,he2024learning}, and imitation~\cite{fu2024humanplus,tang2024humanmimic}. However, current HMR systems are predominantly built on Line-of-Sight (LoS) camera (e.g., RGB and depth) and trained on large-scale video datasets (e.g., Human3.6M~\cite{ionescu2013human3}, 3DPW~\cite{von2018recovering}). This raise two practical issues: (i) visual data exposes personal appearance, often inaccessible for privacy-sensitive scenarios, such as children and elderly care, and public sharing is frequently restricted~\cite{morgan2023multimodal}; (ii) LoS cameras are usually vulnerable to adverse illumination like low-lighting and strong sunlight, and susceptible to occlusion, such as thick clothing or smoke~\cite{chen2022mmbody, ding2025thermohands}.

There is a growing interest in millimeter-wave (mmWave) radar-based human sensing, positioning it as a complement, or even an alternative to vision-based systems~\cite{zhang2023survey, ding2023hidden,zhang2025rf4d,zhou2026tent,zhou2024adapose}. As a radio-frequency (RF) sensor, mmWave radar actively emits radio waves to sense human targets and analyzes the returned echoes to recover spatial information, including range and angles~\cite{ding2022self, rao2017introduction, ding2023hidden}. This sensing mechanism enables trustworthy deployment in complex real-world environments, with robustness to poor illumination and occlusion, while preserving user privacy and facilitating data distribution. Moreover, mmWave radar operates in a higher-frequency band (e.g., 77GHz), offering higher spatial resolution than other RF sensors (e.g., Wi-Fi), making it well-suited for HMR that demands fine-grained observations. 

Despite these advantages, existing large-scale HMR datasets and benchmarks remain largely RGB image-centric, with mmWave-based approaches rarely explored. As summarized in Tab.~\ref{tab:Difference}, most existing mmWave radar-based human sensing datasets~\cite{an2021mars,an2022mri,yang2023mm,zhao2018rf,rahman2024mmvr,ho2024rt} are oriented for coarse-grained HPE and provide only sparse skeleton annotations. These labels are typically derived from RGB(D) cameras via existing pose estimators or multi-view optimization and thus inevitably noisy, introducing ground-truth (GT) bias that hinders high-fidelity sensing. Two RF-based HMR datasets have recently been released~\cite{xue2021mmmesh,chen2022mmbody}, but they remain limited in scale and focus primarily on simple, in-place daily activities, constraining their application to dynamic and diverse scenarios. In terms of the modality, most datasets rely on low-resolution radars that output only very sparse radar point clouds (RPC)~\cite{fuhrmann1992cfar}, discarding much valuable signal information. While recent efforts~\cite{rahman2024mmvr,ho2024rt,ding2024radarocc} have begun to explore raw radar tensors (RT) to preserve more contextual information, they remain centered on coarse-level HPE and do not support fine-grained HMR.

\begin{table*}[!t]
\centering
\caption{Comparison of \sysname with prior datasets ($\dag$ denotes non-public data). Overall, \sysname is the largest RF-based dataset with multi-granularity motion annotations. \textbf{Modalities: }It provides both raw radar tensors (RT) and filtered radar point clouds (RPC) for high-fidelity HMR. \textbf{Annotations:} Human body annotations are obtained with a high-precision marker-based MoCap system (* denotes pseudo-GTs from RGB(D)). \textbf{Diversity: } It extends beyond simple in-place activities to complex, non-in-place rehabilitation and sports. 
}
\renewcommand{\arraystretch}{1}
\setlength\tabcolsep{5pt}
\resizebox{\linewidth}{!}{

    \begin{tabular}{clc|ccc|cccc|cccc|ccc}
    \toprule
    \multirow{2}[2]{*}{{Task}} & \multicolumn{2}{c}{{Dataset}} & \multicolumn{3}{c}{{Modalities}} & \multicolumn{4}{c}{{Annotations}} & \multicolumn{4}{c}{{Statistics}} & \multicolumn{3}{c}{{Actions}} \\
    \cmidrule(lr){2-3} \cmidrule(lr){4-6} \cmidrule(lr){7-10} \cmidrule(lr){11-14} \cmidrule(lr){15-17}
          & \multicolumn{1}{l}{{Name}} & RF-BW(GHz) & RGB   & Depth & mmWave & Actions & Global-Traj. & Skeleton & {Mesh} & \# Act. & \# Sub. & \# Seq. & \# Frame & Daily & Rehab. & Sports \\
    \midrule
    \multirow{3}[2]{*}{\makecell{RGB-based \\ HMR}} & {MPI-INF-3DHP~\cite{mehta2017monocular}} & -     &  \cmark   &       &       &      &    \cmark   &   3D    &  \cmark   & 8    & 8     &   16    & 1.3M   & \cmark     &       &  \\
          & {Human 3.6M}~\cite{ionescu2013human3}  & -     & \cmark     &       &       & \cmark     & \cmark     & 3D    & \cmark     & 17    & 11    & ~210  & 3.6M  & \cmark     &       &  \\
          & {3DPW}~\cite{von2018recovering} & -     &  \cmark    &       &       &     &   \cmark   & 3D    & \cmark     & -     & 7     & 60    & $>$51K  & \cmark     &       &  \\
    \midrule
    \multirow{7}[2]{*}{\makecell{RF-based \\ HPE}} & {mRI}~\cite{an2022mri} & 77-81 & \cmark     & \cmark     & RPC   & \cmark     &       & 3D*   &       & 12    & 20    & 300   & 160K  & \cmark     & \cmark     &  \\
          & {MARS}~\cite{an2021mars} & 77-81 & \cmark     &       & RPC   & \cmark     &       & 3D*   &       & 10    & 4     & 80    & 40K   & \cmark     &       &  \\
          & {RF-Pose3D$\dag$}~\cite{zhao2018rf} & 77-81 &       &       & RT    & \cmark     &       & 3D*   &       & 5     & $>$5    & -     & -     & \cmark     &       &  \\
          & {HrPR}~\cite{lee2023hupr} & 77-81 &       &       & RT    &       &       & 2D*   &       & 3     & 6     & 235   & 141K  & \cmark     &       &  \\
          & {mm-Fi}~\cite{yang2023mm} & 60-64 & \cmark     & \cmark     & RPC   & \cmark     &       & 3D*   &       & 27    & 40    & 1080  & 320K  & \cmark     & \cmark     &  \\
          & {RT-Pose}~\cite{ho2024rt} & 77-81 & \cmark     & \cmark     & RT    &       & \cmark*    & 3D*   &       & 6     & -     & 240   & 72K   & \cmark     &       &  \\
          & {MMVR}~\cite{rahman2024mmvr} & 77-81 & \cmark     &       & RT    &       &     & 2D*   &       & -     & 25    & 395   & 345K  & \cmark     &       &  \\
    \midrule
    \multirow{3}[2]{*}{\makecell{RF-based\\HMR}} & {mmMesh$\dag$}~\cite{xue2021mmmesh} & 77-81 &       &       & RPC   &       &     & 3D    & \cmark     & 8     & 20    & 34    & 3K    & \cmark     &       &  \\
          & {mmBody}~\cite{chen2022mmbody} & 76-81 & \cmark     & \cmark     & RPC   &       &       & 3D    & \cmark     & $<$30 & 20    & 48    & $<$70K & \cmark     &       &  \\
          & \textbf{M4Human (ours)} & {62-69} & {\cmark} & {\cmark} & {RT \& RPC} & \cmark     & \cmark     & 3D    & {\cmark} & {50} & 20    & 999   & 661K  & {\cmark} & {\cmark} & {\cmark} \\
    \bottomrule
    \end{tabular}%
}
\vspace{-1em}
\label{tab:Difference}
\end{table*}

To bridge this gap, we introduce \sysname, the largest-scale multimodal dataset that supports high-fidelity RF-based HMR. It comprises 999 sequences and 661K ($>$15h) synchronized samples across RGB and depth images plus multi-level radar data—raw tensors (RT) and point clouds (RPC)—captured by a high-resolution Vayyar mmWave radar tailored to human sensing. \sysname includes high-quality marker-based motion capture (MoCap) annotations to provide 3D HMR GT, along with an abundant set of meticulously designed daily, rehabilitation, and sports activities. To the best of our knowledge, \sysname is the first and largest benchmark (about $9\times$ larger than the prior largest mmBody~\cite{chen2022mmbody}) to enable mmWave tensor–based HMR across diverse, dynamic, and unconstrained indoor scenarios. Its key features and contributions are summarized as follows:

\begin{itemize}
    \item \textbf{Complementary Sensing Modalities.} \sysname advances multimodal human sensing by integrating complementary sensing modalities, including line-of-sight (LoS) cameras and radio-frequency (RF) mmWave radar. It provides four synchronized modalities: RGB frames, depth frames, raw radar tensors (RT), and radar point clouds (RPC). By including both raw and processed radar data, \sysname supports research across different levels of RF signal processing and representation learning.
     
    \item \textbf{High-Fidelity Motion Annotations.} \sysname includes marker-based motion capture (MoCap) to provide high-fidelity GT annotations, including 3D human meshes, dense pose labels, 2D/3D skeleton keypoints, and full-body trajectories under dynamic free-space sports activities. These annotations, together with abundant action categories, enable fine-grained human sensing research and evaluation.

    \item \textbf{Diverse Action Sets.} 
    \sysname covers 50 categories of human actions, spanning in-place daily activities, clinically guided rehabilitation exercises, and non-in-place sports motions. This diversity broadens the applicability of the dataset to domains such as smart homes, healthcare (e.g., rehabilitation assessment and monitoring), and human-computer interaction (e.g., virtual fitness gaming).

    \item \textbf{Broad Task Support.} 
    The rich annotations and multimodal data in \sysname enable a wide range of research tasks, including human mesh reconstruction (HMR), human tracking, multimodal fusion, and cross-modal supervision for RF-based human sensing. The dataset also supports evaluation under challenging settings such as unseen-subject and unseen-action generalization, facilitating the study of robustness and transferability in real-world deployment scenarios.
    
    \item \textbf{Extensive Benchmarks.} 
    We benchmark state-of-the-art methods for radar-based HMR on both raw radar tensors (RT) and radar point clouds (RPC). We further compare radar-based methods with RGB-D baselines and evaluate multimodal fusion of radar and visual data for improved mesh reconstruction. In addition, we study the effect of dataset scale and demonstrate the utility of radar-based HMR for downstream tasks, including human activity recognition (HAR) and human motion prediction (HMP).
    
\end{itemize}

%
%


%% file: sec/2_related_work.tex
\section{Related Works}

\noindent\textbf{mmWave-based Human Sensing.}
RGB cameras are ubiquitous for human sensing tasks (e.g., HMR~\cite{kocabas2020vibe,dwivedi2024tokenhmr}), but raise privacy concerns in sensitive settings and degrade under occlusion and poor lighting~\cite{chen20232d}. As reported in mmBody~\cite{chen2022mmbody}, camera-based methods suffer remarkable error increases: depth (rain 56.4\%, smoke 336\%), RGB (dark 42.8\%, smoke 16.2\%) on HMR. Commercial mmWave radar has therefore emerged as a privacy-preserving and robust complement for indoor human sensing. Early radar-based works used low-resolution RPCs to tackle coarse-level motion tracking~\cite{zhao2019mid,gu2019mmsense,cui2021high,zhang2024learning}, where human motion was reduced to a single-point trajectory and tracked using clustering and RNNs under simple walking scenarios. Subsequent studies targeted human action recognition for daily actions and HCI gestures~\cite{singh2019radhar,alam2021palmar,liu2020real,li2022towards}. More recently, whole-body keypoint estimation from sparse RPCs has been explored via datasets and methods such as mRI, mm-Fi, and others~\cite{an2021mars,an2022mri,an2022fast,yang2023mm,sengupta2020mm,ding2024milliflow,fan2024diffusion}.  To alleviate the information loss during radar point cloud generation, RETR~\cite{yataka2024retr}, RT-Pose~\cite{ho2024rt} and more~\cite{rahman2024mmvr,wu2022rfmask,lee2023hupr} operate directly on raw radar tensors for human pose estimation. A few recent works, such as mmBody~\cite{chen2022mmbody} and mmMesh~\cite{xue2021mmmesh}, investigate mesh reconstruction from RPCs, but they remain limited to daily actions and in data scale, which limits their application scope and generaliability. 
To date, no works have explored RT-based HMR. 
 
\noindent\textbf{mmWave-based Human Pose Datasets.}
The success of RGB-based HMR methods has been driven by large-scale image-based datasets (e.g., MPI-INF-3DHP~\cite{mehta2017monocular}, Human3.6M~\cite{ionescu2013human3}, 3DPW~\cite{von2018recovering}), which provide high-quality 3D annotations from marker-based or markerless multi-view MoCap systems. In contrast, early mmWave HPE datasets~\cite{an2021mars,zhao2018rf} focus on simple in-place daily activities and provide only sparse RPC from low-resolution (LR) radars (e.g., 4 Tx/Rx channels, 4 GHz bandwidth; see Table~\ref{tab:Difference}). mRI~\cite{an2022mri} and mm-Fi~\cite{yang2023mm} expand to in-place rehabilitation movements but still rely on synchronized RGB to annotate 3D skeletons, which is inevitably noisy and hard to support high-fidelity human sensing. More recent datasets~\cite{lee2023hupr,wu2022rfmask}, such as MMVR~\cite{rahman2024mmvr}, RT-Pose~\cite{ho2024rt}, explore raw radar tensors (RT) for coarse-grained HPE, often using cascaded LR radar chips or multi-view setups to boost spatial resolution at the cost of harder synchronization and calibration. For HMR, mmMesh~\cite{xue2021mmmesh} first reconstructs MoCap-annotated meshes from LR RPC, but its dataset is not publicly available. mmBody~\cite{chen2022mmbody}, employs a high-resolution automotive radar for indoor perception but lacks access to more informative RT data. Moreover, its RPC contains substantial background returns, leaving only a small fraction from the human foreground (c.f. Fig.~\ref{vsmmbody}(a)). In addition, existing radar datasets remain limited in scale and largely restricted to simple, in-place actions, which constrains their utility for broader and more dynamic human activities (c.f. Fig.~\ref{vsmmbody}(b)).

%% file: sec/3_method.tex
\section{Datasets}

\begin{figure*}[!ht]
\centering
\includegraphics[width=.9\linewidth]{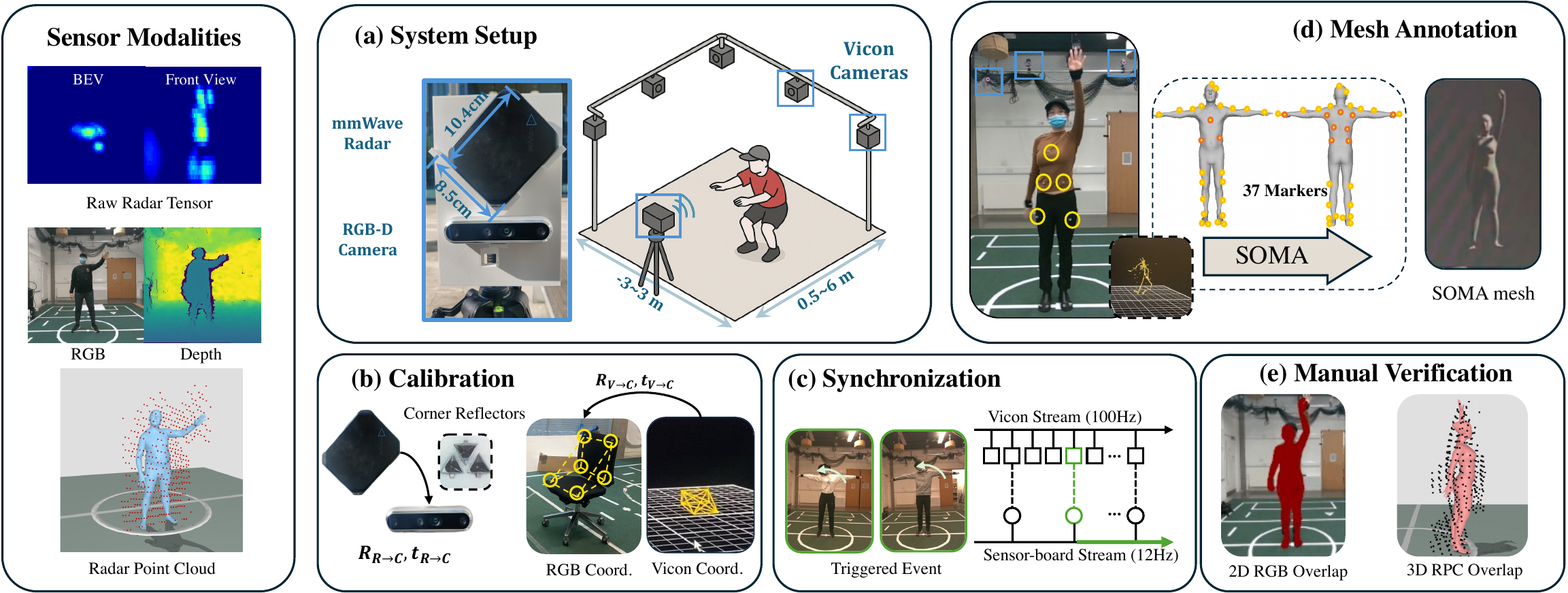}
\caption{Overview of the system setup. \sysname designs a multimodal sensing platform with high-precision marker-based MoCap system. Appropriate calibration and synchronization workflow are designed for accurate alignment between modalities and annotations.} 
\vspace{-1em}
\label{systemsetup}
\end{figure*}

\noindent\textbf{Overview.}
At the core of our contribution is \sysname, a large-scale multimodal dataset designed for fine-grained human mesh reconstruction (HMR). As illustrated in Fig.~\ref{systemsetup}, we build a unified sensing platform that integrates a high-resolution mmWave radar, an RGB–D camera, and a Vicon MoCap system. During capture, subjects are instructed to perform a diverse set of actions while the RGB-D camera and radar record synchronized multimodal data, and the Vicon system captures precise 3D marker positions for accurate mesh annotations. Our dataset is characterized by (i) \emph{diversity}: 661K frames across 999 sequences, involving 20 subjects and 50 meticulously designed action classes; (ii) \emph{multimodality}: synchronized RGB, depth, radar point cloud (RPC), and raw radar tensor (RT) with precise cross-modal calibration; (iii) \emph{high-quality annotation}: dense and time-aligned 3D meshes derived from marker-based Vicon capture with human-in-the-loop verification.

\begin{figure}[!t]
\centering
\includegraphics[width=.95\linewidth]{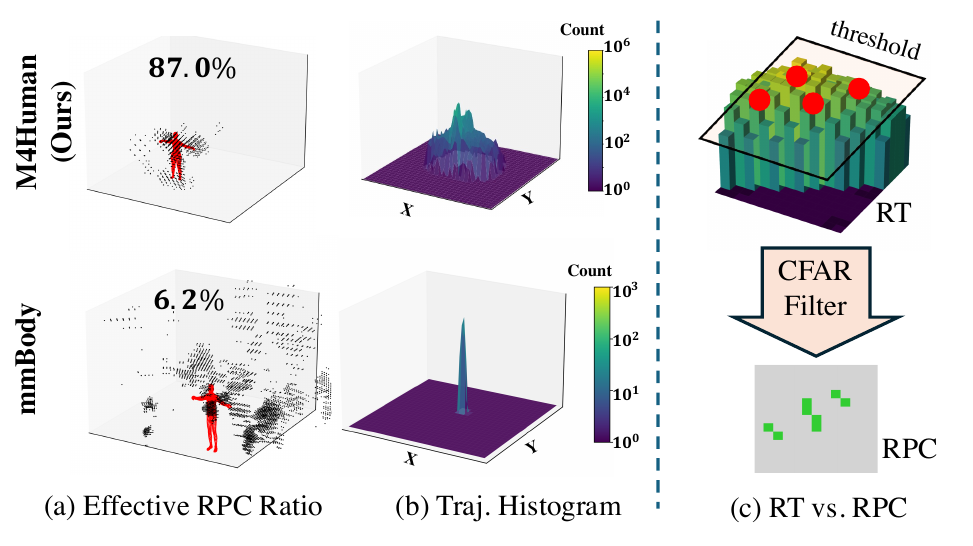}
\caption{Comparison between \sysname and mmBody~\cite{chen2022mmbody}: (a) \sysname achieves a much higher effective RPC ratio around human subjects. (b) \sysname is much larger in scale and involves more diverse global trajectories than simple in-place daily activities of mmBody. (c) \sysname provides additional raw RT modality that contains more unfiltered information than RPC.}
\vspace{-1em}
\label{vsmmbody}
\end{figure}

\subsection{System Setup} 

We use an Intel RealSense D435 stereo depth camera~\cite{IntelRealSense2023D455} that integrates an IR projector, an RGB camera, and a stereo IR pair to acquire synchronized RGB–D image frames. Raw resolution is up to 1280$\times$720; for storage efficiency and faster I/O, we downsample to 640$\times$480. mmWave data are captured using the commercial Vayyar vTrigB imaging mmWave radar~\cite{vayyarhome}. The device integrates a proprietary human-sensing mmWave Software on Chip (SoC) and outputs both raw radar tensors (RT) and processed radar point clouds (RPC). With 7 GHz bandwidth and a large 20 Tx/Rx antenna array, it provides markedly better range and angular resolution than common 3-4 Tx/Rx TI radars~\cite{rao2017introduction}. The mmWave radar and RGB–D camera are mounted on a 3D-printed bracket fixed to a tripod (c.f. Fig.~\ref{systemsetup} (a)), connected to a unified recording PC. Subjects are recorded at distances of 0.5–6 m from the platform, within a motion area of approximately $6 \times 5.5\mathrm{m}^2$, with multiple facing orientations for data diversity. A Vicon MoCap system is installed overhead, providing high-precision position measurements of 3D markers attached to anatomical keypoints during action execution.



 




\subsection{Human-centric and Multi-level Radar Data} 

The Vayyar radar used in \sysname is tailored to fine-grained human sensing. To show its superior sensing ability, we define an \textit{effective RPC ratio} metric as the fraction of points near humans relative to all points per frame (c.f. Fig.~\ref{vsmmbody}(a)). \sysname attains a substantially higher effective RPC ratio than mmBody~\cite{chen2022mmbody}, which employed automotive radars designed for broad scene sensing rather than human targets, yielding much more background targets. With more human-concentrated returns, \sysname supports simultaneous global tracking and high-fidelity mesh reconstruction for diverse, non-in-place dynamic activities in open spaces. Furthermore, \sysname provides two complementary radar modalities: raw radar tensor (RT) and radar point cloud (RPC). RT is a 3D intensity volume obtained via FFT-based processing of time-domain signals across range, azimuth, and elevation axes, then mapped into Cartesian coordinates (X-Y-Z). RPC is derived from RT using CFAR~\cite{fuhrmann1992cfar}, which retains only salient reflections above adaptive thresholds. As illustrated in Fig.~\ref{vsmmbody}(c), CFAR removes sub-threshold structures and suppresses much of the spatial intensity context. Therefore, RT generally contains richer information than RPC. To support clearer interpretation and more comprehensive learning from mmWave signals, \sysname releases both RT and RPC modalities for public usage.


\subsection{Calibration}

For spatial calibration between the Vicon system and sensor platform, we affix six non-coplanar MoCap markers to a chair placed in the motion area (Fig.~\ref{systemsetup} (b)). We record their 3D positions in the Vicon frame \(\mathbf{X}_i^V\) ($i\in[1,..,6]$) and manually annotate the corresponding 2D pixel locations $\mathbf{x}_i$ in the RGB image. With chessboard-calibrated camera intrinsics $\mathbf{K} \in \mathbb{R}^{3\times3}$, we solve a Perspective-\(n\)-Point (PnP) problem~\cite{fischler1981random} to estimate the camera extrinsics \((\mathbf{R}_{C \leftarrow V}, \mathbf{t}_{C \leftarrow V})\), such that \(\mathbf{x}_i \sim \mathbf{K}(\mathbf{R}_{C \leftarrow V}\mathbf{X}_i^V+\mathbf{t}_{C \leftarrow V})\). We obtain the radar–camera extrinsic calibration \((\mathbf{R}_{C \leftarrow R}, \mathbf{t}_{C \leftarrow R})\) following a similar procedure but using radar-visible targets (e.g., corner reflectors) instead of MoCap markers. With both transformations, any 3D Vicon point \(\mathbf{P}^V\) can be first transformed into the camera frame as \(\mathbf{P}^C=\mathbf{R}_{C \leftarrow V}\mathbf{P}^V+\mathbf{t}_{C \leftarrow V}\), and then to the radar frame via \(\mathbf{P}^R=\mathbf{R}_{C \leftarrow R}^{-1}(\mathbf{P}^C-\mathbf{t}_{C \leftarrow R})\). This calibration process yields precise cross-modal alignment among the Vicon, RGB–D camera, and radar data.

\subsection{Time Synchronization} 

In our setup, the Vicon system and the sensor board run on seperate PCs. Radar and RGB-D streams are polled by one PC, achieving synchronization at the maximum RT transfer rate of 12Hz. To synchronize with Vicon, subjects are instructed to perform a trigger gesture when recording starts: after a T-pose, they rapidly swing their head to the right (c.f. Fig.~\ref{systemsetup} (c)). We detect the Vicon start frame as the first head-top marker’s displacement exceeds 10 cm, and we manually locate the corresponding RGB start frame and trim earlier radar/RGB–D frames accordingly. Because the Vicon stream has a higher frame rate (100Hz), each sensor frame is then temporally matched to its nearest Vicon frame. 

\subsection{Subjects and Actions}

\noindent\textbf{Subjects.} \sysname dataset features a high diversity of human subjects, comprising 20 volunteers with varied heights (1.5–1.8m) and weights (45–90kg), balanced genders (12 male, 8 female), and heterogeneous ethnic backgrounds. This wide range of subject attributes ensures the dataset’s versatility and generaliability to a broad population. All participants have been informed that the data will be made publicly available for research purposes, and were instructed to wear caps and masks for de-identification.

\noindent\textbf{Actions.} \sysname meticulously curates 50 actions spanning three categories: (i) \textit{daily activities} (e.g., waving, walking in a curve, sit-and-stand), targeting smart-home applications and human–computer interaction; (ii) \textit{rehabilitation exercises} (e.g., upper limb extension, high knees) motivated by clinical practice to support routine training~\cite{zhao2024enabling} and recovery assessment~\cite{kidzinski2020deep} in healthcare applications; and (iii) non-in-place \textit{sports actions} (e.g., basketball, badminton) that require substantial whole-body coordination, supporting dynamic motion analysis and VR/AR applications.


During capture, each subject performs all 50 actions, with at least 550 frames captured per action at 12 Hz. This led to a total of 999 valid sequences, $\sim$661K valid synchronized multimodal frames, and $\sim$15.3 hours of motion capture. 



\begin{figure*}[!ht]
\centering
\includegraphics[width=.9\linewidth]{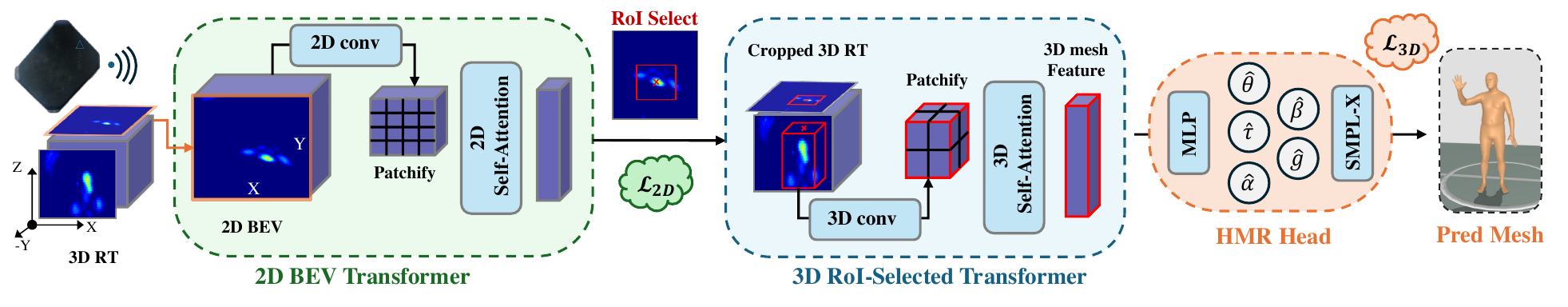}
\caption{Overview of the proposed RT-Mesh baseline. Given a 3D radar tensor (RT), RT-Mesh first reshapes it into a 2D BEV representation. A lightweight 2D BEV Transformer, combining 2D convolution and self-attention, performs efficient 2D human localization $(\hat{x},\hat{y})$ under the supervision of $\mathcal{L}_{2D}$. A local 3D RoI is cropped from the full RT volume based on $(\hat{x},\hat{y})$, which is then processed by 3D convolution and 3D Transformer to extract fine-grained 3D mesh features. Finally, an HMR head regresses SMPL-X parameters for 3D mesh.}
\vspace{-1.em}
\label{baseline_fig}
\end{figure*}

\subsection{Body Mesh Annotation}
\label{mesh_annotations}
We obtain high-precision human body mesh annotations via a three-stage pipeline. (i) \emph{Marker-based capture}. We record human motion with a Vicon MoCap system that tracks the subject as a non-rigid body parameterized by 37 anatomically distributed markers (Fig.~\ref{systemsetup} (d)). This yields accurate 3D whole-body trajectories and stable root motion, providing higher-fidelity labels than the purely image-based annotation approaches common in prior RF datasets. (ii) \emph{Human-in-the-loop cleanup}. The automatically labeled Vicon sequences are inspected frame by frame to correct typical MoCap errors (marker swaps, short-term occlusions, missing markers), so that the final marker trajectories are temporally consistent, stable, and avoid drifting. (iii) \emph{Mesh reconstruction and validation}. Cleaned marker trajectories are fed to SOMA~\cite{ghorbani2021soma}, a neural marker-to-body reconstructor, to estimate full-body pose, global trajectories, and SMPL-X-style meshes~\cite{pavlakos2019expressive}. The reconstructed meshes and joints are then visualized together with the RGB-D and mmWave data to verify spatial-temporal alignment (Fig.~\ref{systemsetup} (e)).

%% file: sec/4_experiment.tex
\section{Benchmark and Evaluation}

\paragraph{Overview.} \sysname enables a comprehensive benchmark across sensing modalities and tasks. We evaluate mmWave-based HMR with two radar representations, RPC and RT (Sec.~\ref{sec:radar-based hmr}); comparing radar- and image-based HMR to quantify each modality’s strengths and remaining gaps (Sec.~\ref{sec:rgbd}); and exploring modality fusion to show how mmWave radar benefits multimodal HMR (Sec.~\ref{sec:multimodal}). Since no prior works have used RTs for HMR, we propose a simple baseline supporting the RT modality (Sec.~\ref{baseline:rt-mesh}). We further demonstrate that the scale of \sysname improves the generalizability of radar-based HMR (Sec.~\ref{sec:in-depth radar}), and that radar-predicted human meshes support downstream tasks such as human action recognition (HAR)  (Sec.~\ref{sec:har}) and generative human motion prediction (HMP)~\cite{fan2025mmpred} (in Supplementary).

\subsection{Benchmark Setup}

\noindent\textbf{Protocol.} We define three evaluation protocols aligned with action groups to assess performance under varying difficulty levels. Protocol 1 (P1) covers 30 \emph{in-place} daily and rehabilitation activities. Protocol 2 (P2) focuses on five \emph{sit-in-place} daily and rehabilitation exercises. Protocol 3 (P3) comprises all \emph{non-in-place} daily/sports activities, which are more challenging due to larger displacements and rapid motion. We also report results on the full set (ALL).

\noindent\textbf{Split.} 
The dataset is divided into train/val/test with a ratio of 75:5:20, yielding $\sim$496K/34K/132K samples. We evaluate under three split settings: (S1) Random, all clips are randomly assigned to non-overlapping subsets, modeling seen-subject and seen-action cases; (S2) Cross-Subject, no subject in training appears in testing (two males and two females), modeling unseen-subject generalization; (S3) Cross-Action, a \emph{group-stratified} split with disjoint action classes; the test set includes 20\% of classes sampled from each action group (i.e., in-place, sit-in-place, non-in-place), assessing unseen-action generalization.

\noindent\textbf{Metrics.}
We evaluate the mesh reconstruction performance using four classic metrics, all computed in the \emph{world} frame without root/Procrustes alignment and reported in millimeters/degrees. \textit{Mean Vertex Error (MVE)} is the average Euclidean distance between predicted and ground-truth SMPL-X vertices over all 10,475 vertices. \textit{Mean Joint Error (MJE/MPJPE)} is the mean Euclidean distance between corresponding 3D joints using the 22-joint SMPL-X set. \textit{Mean Rotation Error (MRE)} is the geodesic angular error on $SO(3)$ between predicted and ground-truth joint rotations, averaged across joints; it emphasizes pose and is invariant to global translation. \textit{Translation Error (TE)} is the Euclidean distance between predicted and ground-truth global root translations.

\begin{table*}[!t]
\centering
\caption{Performance of SOTA radar-based HMR using RPC and RT modalities. The mean vertex error (MVE) (mm) is recorded for all protocols and splits, lower the better. We also include the single-sample Latency (Lat.) and GFLOPs for comparing model efficiency.}
\renewcommand{\arraystretch}{1}
\setlength\tabcolsep{8pt}
\resizebox{\linewidth}{!}{

    \begin{tabular}{clcc|rrrrrrrrrrrr}
    \toprule
    \multirow{2}[2]{*}{\textbf{Modality}} & \multicolumn{1}{c}{\multirow{2}[2]{*}{\textbf{Methods}}} & \multicolumn{2}{c}{\textbf{Efficiency}} & \multicolumn{3}{c}{\textbf{P1 (In-Place)}} & \multicolumn{3}{c}{\textbf{P2 (Sit-In-Place)}} & \multicolumn{3}{c}{\textbf{P3 (Non-In-Place )}} & \multicolumn{3}{c}{\textbf{ALL}} \\
    \cmidrule(lr){3-4}  \cmidrule(lr){5-7}  \cmidrule(lr){8-10} \cmidrule(lr){11-13} \cmidrule(lr){14-16}   
          &       & \textbf{Lat. (ms)} & \textbf{GFLOPs} & \multicolumn{1}{c}{\textbf{S1}} & \multicolumn{1}{c}{\textbf{S2}} & \multicolumn{1}{c}{\textbf{S3}} & \multicolumn{1}{c}{\textbf{S1}} & \multicolumn{1}{c}{\textbf{S2}} & \multicolumn{1}{c}{\textbf{S3}} & \multicolumn{1}{c}{\textbf{S1}} & \multicolumn{1}{c}{\textbf{S2}} & \multicolumn{1}{c}{\textbf{S3}} & \multicolumn{1}{c}{\textbf{S1}} & \multicolumn{1}{c}{\textbf{S2}} & \multicolumn{1}{c}{\textbf{S3}} \\
    \midrule
    \multirow{2}[2]{*}{RPC} & mm-Mesh~\cite{yang2023mm} &  3.53     &  2.87     &   105.9    &    149.4   &   146.0    &    183.3   &   202.8    &   194.5    &  201.4     &   226.8    &   223.3    &   132.7    &  170.1     & 173.8 \\
          & P4Trans.~\cite{chen2022mmbody} &   7.17    &   11.76    & 72.7  & 129.3 & 132.3 & \textbf{115.2} & 142.8 & 132.0 & \textbf{139.7} & 180.2 & 184.4 & \textbf{90.4} & 140.8 & 147.8 \\
    \midrule
    \multirow{3}[2]{*}{RT} & RT-Pose~\cite{ho2024rt} &   39.58    &   50.67    & 80.0  & 135.9 & 133.7 & 130.2 & 152.4 & 139.5 & 158.5 & 188.4 & 196.0 & 100.7 & 148.1 & 152.8 \\
          & RETR~\cite{yataka2024retr}  &   17.87    &   3.01    & 73.2  & 159.4 & 143.0 & 133.6 & 169.4 & 153.6 & 162.6 & 206.1 & 207.0 & 97.1  & 169.7 & 163.1 \\
          & RT-Mesh (ours) &  \textbf{2.74}     &   \textbf{2.60}    & \textbf{72.4} & \textbf{123.6} & \textbf{128.5} & 118.1 & \textbf{138.5} & \textbf{126.5} & 142.0 & \textbf{173.6} & \textbf{178.2} & 90.9  & \textbf{135.1} & \textbf{143.1} \\
    \bottomrule
    \end{tabular}%

    }
\label{tab:radar-hmr}
\end{table*}

\subsection{RT-Mesh: A Baseline for RT-based HMR}
\label{baseline:rt-mesh}

\noindent\textbf{Problem Formulation.}
Following prior works~\cite{yang2023mm,an2022fast,an2022mri}, we allow a short temporal window of radar frames. We stack $T$ = $4$ consecutive RTs (3 historical frames), forming a 4D tensor $X_\text{RT} \in \mathbb{R}^{T \times X  \times Y \times Z }$, where $(X,Y,Z)=(121,111,31)$ in our dataset. Given $X_\text{RT} $, our goal is to regress the current-frame SMPL-X parameters $(\alpha,\beta, \tau, \theta)$: root orientation (axis–angle) $\alpha \in \mathbb{R}^3$, body shape $\beta \in \mathbb{R}^{10}$, global translation $\tau \in \mathbb{R}^3$, and body pose $\theta \in \mathbb{R}^{22\times 3}$. We also regress gender probability $g \in [0,1]$ to select the appropriate male/female SMPL-X template. Supervision for $(\alpha,\beta, \tau, \theta, g)$ comes from our MoCap-derived mesh annotations (Sec.~\ref{mesh_annotations}).

\noindent\textbf{Model Architecture.} We propose \textbf{RT-Mesh}, the first HMR method that operates directly on RTs. 
Compared with RPC, RT retains richer human return signals but at much higher volume, making full-volume 3D/4D convolutions computationally prohibitive. To improve the model efficiency with losing key context information, RT-Mesh is specifically designed as a two-stage structure (c.f. Fig.~\ref{baseline_fig}): it first localizes the human foreground RoI amid clutter using a collapsed 2D BEV tensor, then operates on a local 3D RT crop to regress the final SMPL-X mesh parameters. Detailed architecture and training loss designs are provided in the supplementary.




\subsection{Competing Methods}
\label{sec:competing}
\noindent\textbf{RGB(D) Modality.}
For the RGB setting, we adopt the SOTA TokenHMR method~\cite{dwivedi2024tokenhmr} and follow its practices: persons are first detected using Detectron2~\cite{wu2019detectron2} and cropped for local mesh recovery. To obtain the best performance, we initialize the model with its large-scale pretrained weights and fine-tune it on our dataset. For the depth setting, we back-project depth pixels to 3D point clouds and apply the point cloud-based HMR method P4Transformer~\cite{fan2021point}.

\noindent\textbf{mmWave Modality.}
To comprehensively benchmark radar-based HMR, we include SOTA methods from existing mmWave HPE and HMR datasets for both RPC and RT modalities. For RPC-based methods, we select two representative HMR approaches: LSTM-based mmMesh~\cite{xue2021mmmesh} and transformer-based P4Transformer~\cite{fan2021point} applied in the mmBody~\cite{chen2022mmbody} dataset. Since no existing HMR method is designed for RT, we adopt SOTA models from the latest RT-based HPE datasets: MMVR~\cite{rahman2024mmvr} and RT-Pose~\cite{ho2024rt}. Specifically, RT-Pose~\cite{ho2024rt} adapts HR-Net to radar, using computationally expensive 3D CNN network as backbone. We adapt RETR~\cite{yataka2024retr} from MMVR to extract features from two complementary views (X-Y BEV and X-Z side view), using a transformer-based architecture. We preserve the original RT feature extractors, replace their HPE heads with HMR heads, and retrain the models on our dataset.

%% file: sec/5_discussion.tex
\section{Results and Discussion}

\subsection{Radar-based HMR Benchmark}
\label{sec:radar-based hmr}

Table~\ref{tab:radar-hmr} compares radar-only HMR methods across two radar modalities, RPC and RT. Under the random split (S1), both modalities performs strongly: P4Trans. (RPC)~\cite{fan2021point} and RT-Mesh (RT) reach $\sim$70mm MVE on In-Place actions and $\sim$90mm overall, exceeding reported results on prior radar-based HMR/HPE datasets, e.g., mmBody~\cite{chen2022mmbody}. This underscores the quality of our dataset and shows that high‑resolution mmWave radar alone can support accurate HMR in privacy‑sensitive or vision‑impaired scenarios. Our dataset poses additional challenges for another two protocols and split settings, and resulted in performance drop. P2 (Sit-In-Place) suffers from frequent self-occlusion and chair-induced multipath noise, while P3 (Non-In-Place) require jointly tracking and pose/mesh estimation under rapid, dynamic motion. S2 (Cross-subject) and S3 (Cross-action) settings also remain challenging, highlighting the need for better generaliability.

\begin{figure*}[!ht]
\centering
\includegraphics[width=0.95\linewidth]{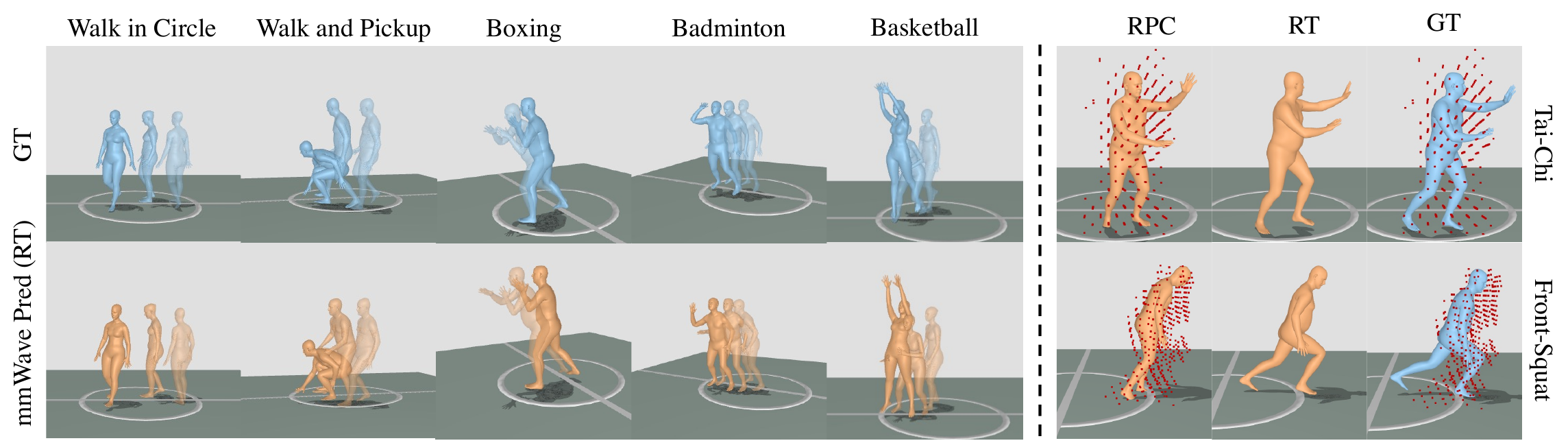}
\caption{Visualization of (left) RT-based HMR under challenging (P3) non-in-place actions, and (right) comparison between RPC and RT predicted meshes. The proposed RT-Mesh can simultaneously track and reconstruct 3D human meshes during complex sports motions. In contrast, RPC-based prediction may fail when points are missing for certain body parts. 
}
\label{vis}
\end{figure*}

Comparing two radar representations, RT and RPC perform similarly on S1, but RT outperforms RPC on S2 and S3, indicating stronger generaliability. RT preserves denser, more continuous spatial evidence, whereas RPC is sparse and susceptible to missed detections of body parts (c.f. Fig.~\ref{vis}). Especially for novel subjects and actions, these gaps can induce overfitting and yield twisted meshes. Moreover, our proposed RT-Mesh attains markedly higher efficiency than existing RPC- and RT-based methods, particularly those that process full-volume RT. It achieves 2.74 ms latency and 2.6 GFLOPs, making it deployable on edge devices. As a simple and efficient baseline, RT-Mesh is designed to encourage exploration into RT-based HMR, which remains an underexplored direction. We believe future work can build on this foundation to develop models with higher spatial precision, temporal consistency, and integration of human priors.

\begin{figure}[!b]
\centering
\vspace{-2em}
\includegraphics[width=\linewidth]{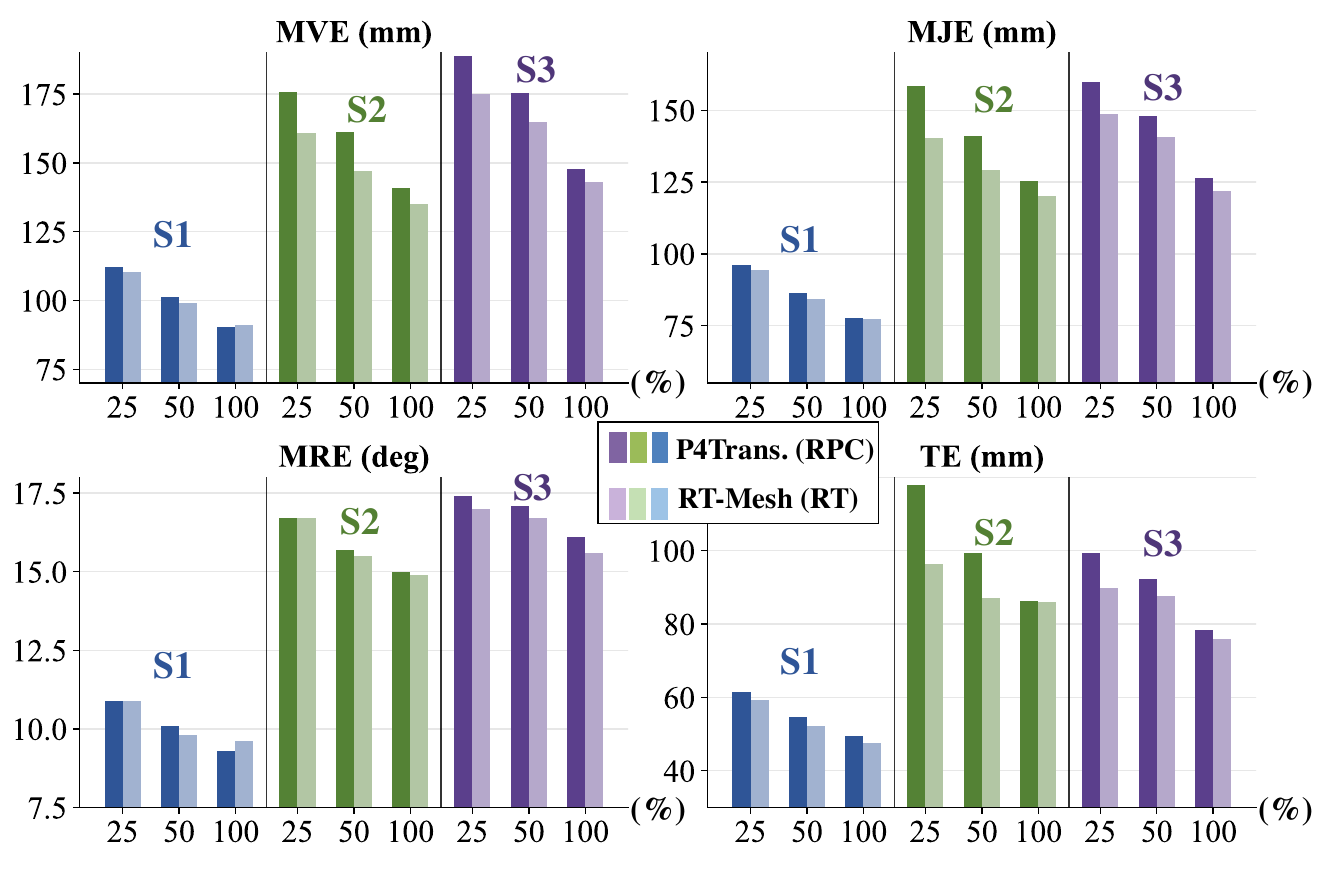}
\caption{Impact of the training dataset size on radar-based HMR. Larger dataset consistently improves performance on S2 (cross-subject) and S3 (cross-action) across all evaluation metrics.
}
\label{tab:scale}
\end{figure}

\begin{table*}[!t]
\centering
\caption{Performance of different single modalities and multi-modality fusion under protocol ALL and 3 different splits. All four metrics are reported to reflect the advantages of different modalities, lower the better. 
}
\renewcommand{\arraystretch}{1}
\setlength\tabcolsep{10pt}
\resizebox{\linewidth}{!}{

    \begin{tabular}{cc|c|cccc|rrrr|rrrr}
    \toprule
    \multicolumn{2}{c|}{\multirow{2}[2]{*}{\textbf{Modality}}} & \multirow{2}[2]{*}{\textbf{Protocol}} & \multicolumn{4}{c|}{\textbf{S1 (Random)}} & \multicolumn{4}{c|}{\textbf{S2 (Cross-Sub)}} & \multicolumn{4}{c}{\textbf{S3 (Cross-Act)}} \\
    \multicolumn{2}{c|}{} &       & \textbf{MVE} & \textbf{MJE} & \textbf{MRE} & \textbf{TE} & \multicolumn{1}{c}{\textbf{MVE}} & \multicolumn{1}{c}{\textbf{MJE}} & \multicolumn{1}{c}{\textbf{MRE}} & \multicolumn{1}{c|}{\textbf{TE}} & \multicolumn{1}{c}{\textbf{MVE}} & \multicolumn{1}{c}{\textbf{MJE}} & \multicolumn{1}{c}{\textbf{MRE}} & \multicolumn{1}{c}{\textbf{TE}} \\
    \midrule
    \multirow{4}[2]{*}{\textbf{Single}} & RGB   & \multirow{7}[6]{*}{ALL} & 97.5  & 87.0  & 7.4   & 54.4  & \multicolumn{1}{c}{149.7} & \multicolumn{1}{c}{128.4} & \multicolumn{1}{c}{9.3} & \multicolumn{1}{c|}{99.2} & \multicolumn{1}{c}{116.7} & \multicolumn{1}{c}{110.6} & \multicolumn{1}{c}{10.0} & \multicolumn{1}{c}{82.7} \\
          & Depth  &       & 82.7  & 72.1  & 9.1   & 45.6  & \multicolumn{1}{c}{127.1} & \multicolumn{1}{c}{115.5} & \multicolumn{1}{c}{14.4} & \multicolumn{1}{c|}{89.1} & \multicolumn{1}{c}{123.2} & \multicolumn{1}{c}{106.9} & \multicolumn{1}{c}{14.4} & \multicolumn{1}{c}{61.2} \\
         & RPC   &       & 90.4  & 77.5  & 9.3   & 49.4  & \multicolumn{1}{c}{140.8} & \multicolumn{1}{c}{125.2} & \multicolumn{1}{c}{15.0} & \multicolumn{1}{c|}{86.4} & \multicolumn{1}{c}{147.8} & \multicolumn{1}{c}{126.4} & \multicolumn{1}{c}{16.1} & \multicolumn{1}{c}{78.4} \\
          & RT    &       & 90.9  & 77.2  & 9.6   & 47.6  & \multicolumn{1}{c}{135.1} & \multicolumn{1}{c}{120.2} & \multicolumn{1}{c}{14.9} & \multicolumn{1}{c|}{86.1} & \multicolumn{1}{c}{143.1} & \multicolumn{1}{c}{122.0} & \multicolumn{1}{c}{15.6} & \multicolumn{1}{c}{76.0} \\
\cmidrule{1-2}\cmidrule{4-15}    \multirow{3}[2]{*}{\textbf{Fusion}} & RPC + RT &       & 84.3  & 71.7  & 8.9   & 43.8  &    135.2   &   119.3    &   15.0    &  85.7     &   140.8    &   120.9    &    15.5   & 73.4 \\
          & RGB + RT &       & 80.1  & 71.4  & 8.9   & 51.6  &   \textbf{112.5}    &   103.2    &  \textbf{11.6}     &  80.5     &  \textbf{108.7}     &   \textbf{98.4 }   &  \textbf{11.5}     & 75.7 \\
          & Depth + RT &       & \textbf{77.5} & \textbf{66.3} & \textbf{8.4} & \textbf{36.0} &   115.9    &   \textbf{102.6}    &   14.0    &   \textbf{72.2}    &     120.0  &  102.4     &     14.6  & \textbf{56.6} \\
    \bottomrule
    \end{tabular}%
}
\label{tab:modality}
\end{table*}

\begin{table}[!t]
\centering
\caption{Benchmark on downstream skeleton-based human action recognition (HAR) out of 50 actions. We show the Top-1 and Top-5 accuracy (\%) using GT skeletons on S1, radar tensor (RT) predicted skeletons on S1 and S2.}

\renewcommand{\arraystretch}{1}
\setlength\tabcolsep{8pt}
\resizebox{\linewidth}{!}{

    \begin{tabular}{l|cccccc}
    \toprule
    \multirow{2}[2]{*}{Methods} & \multicolumn{2}{c}{GT (S1)} & \multicolumn{2}{c}{RT (S1)} & \multicolumn{2}{c}{RT (S2)} \\
          & Top-1 & Top-5 & Top-1 & Top-5 & Top-1 & Top-5 \\
    \midrule
    CNN~\cite{du2015skeleton}   & 53.55 & 80.76 & 37.73 & 51.37 & 41.26 & 51.14 \\
    AGCN~\cite{shi2019two}  & 65.70  & 94.24 & \textbf{64.82} & \textbf{90.55} & \textbf{59.91} &\textbf{ 83.97} \\
    BlockGCN~\cite{zhou2024blockgcn} & \textbf{75.67} & \textbf{95.22} & 61.06 & 87.75 & 27.85 & 51.01 \\
    \bottomrule
    \end{tabular}%
}
\vspace{-1em}
\label{tab:har}
\end{table}

\subsection{In Depth Analysis of Dataset Scale}
\label{sec:in-depth radar}
Fig.~\ref{tab:scale} presents an in-depth analysis of how the  training dataset size influences the performance of radar-based HMR. We evaluate models trained with 25\%, 50\%, and 100\% of our training data under different split settings, where 25\% approximates the size of existing radar HMR datasets such as mmBody~\cite{chen2022mmbody}. For cross-subject (S2), this corresponds to using 4, 8, and 16 training subjects; for cross-action (S3), we use 10, 20, and 40 training action classes, respectively. 

We find that the scaling law~\cite{kaplan2020scaling} holds for mmWave-based HMR: enlarging the training set consistently improves performance across splits and metrics, particularly for S2 and S3. For instance, RT-Mesh improves from 161.0mm to 135.1mm MVE on S2, and from 174.9mm to 143.1mm on S3. A richer dataset provides essential diversity for learning robust mappings between human motion and radar reflections, which helps mitigate the common challenges of multi-path interference and body-part miss-detection inherent to RF sensing. These results demonstrates the importance of large-scale training data in advancing mmWave-based human sensing and modelling, potentially enabling hard use cases (e.g., HMR for sports) and narrowing the gap to RGB-D systems.

\subsection{Comparison to RGB(D)-based Methods}
\label{sec:rgbd}
Table~\ref{tab:modality} (Single) compares the best RPC- and RT-based model against RGB(D)-based methods (c.f. Sec.~\ref{sec:competing}). With higher-resolution radar, radar-only models achieve comparable performance to RGB-D. Under S1 and S2, radar surpasses RGB and approaches depth. Notably, the translation error (TE) of radar matches depth and clearly outperforms RGB, which lacks explicit range information. We attribute this to radar’s higher sensitivity to moving foreground and suppression of static background, yielding more reliable root tracking in fast, non-in-place motions. Under S2 and S3, all modalities degrades largely, indicating that generalization to unseen subjects/actions remains difficult, even for camera-based HMR models that benefit from large-scale pretraining. We expect \sysname to serve as a strong benchmark for advancing generalizable HMR across all modalities.

\subsection{Multimodal Fusion}
\label{sec:multimodal}
Table~\ref{tab:modality} (Fusion) benchmarks multi-modal fusion for HMR. For this experiment, we apply intermediate feature fusion, where features from each modality are concatenated before the HMR prediction head. On S1 and S3, RPC+RT remarkably outperforms either RPC or RT alone. This result highlights the complementary value of RT and underscores the significance of \sysname as the first dataset supporting RT-based HMR. We argue that learning jointly from two radar representations is mutually reinforcing: RT preserves dense spatial context while RPC emphasizes salient targets with pronounced motion. Moreover, fusing RT with RGB and depth images also yield significant gains, especially in TE for root-trajectory tracking. This demonstrates that radar can be a strong complementary modality to many camera-based systems for real-world applications.

\subsection{Downstream HAR Benchmark}
\label{sec:har}
Table~\ref{tab:har} evaluates skeleton-based HAR with two inputs: (i) MoCap-de skeletons GT and (ii) RT-estimated skeletons extracted from meshes predicted by RT-Mesh. Under S1 (random split), BlockGCN~\cite{zhou2024blockgcn} attains $\sim$75\% Top-1 accuracy using MoCap skeletons, highlighting that our dataset is challenging due to dynamic, non–in-place actions. Moreover, RT-estimated skeletons already exhibit considerable accuracy, reaching 64.82\% (Top-1), indicating that the proposed RT-Mesh can effectively support downstream HAR. Cross-subject generalization is still challenging for radar-based HAR, given additional performance drops in S2. 

%% file: sec/6_conclusion.tex
\section{Conclusion}
We present \sysname, a large-scale multimodal benchmark for mmWave radar–based human mesh reconstruction. The dataset comprises 661K synchronized frames spanning 50 actions and 20 subjects, captured with high-resolution mmWave radar, RGB, and depth modalities, and annotated with marker-based 3D meshes. \sysname provides both raw radar tensors (RT) and radar point clouds (RPC) to enabling research across different RF representation levels. We also propose RT-Mesh, a simple but effective baseline—the first method to perform HMR directly from raw RTs. Extensive experiments establish single- and multi-modal benchmarks across radar and RGB-D modalities, revealing the potential of mmWave radar for fine-grained, privacy-preserving human sensing. We expect \sysname to serve as a foundation for advancing RF-based 3D human modeling and its applications in Physical AI systems.

\vspace{1em}
\small{
\noindent\textbf{Acknowledgment.}}
This research is partially supported by EPSRC under the Centre for Doctoral Training in Robotics and Autonomous Systems at the Edinburgh Centre of Robotics (EP/S023208/1). This work is supported by Ministry of Education (MOE), Singapore, under AcRF TIER 1 Grant RG64/23. This work is jointly supported by MOE Singapore Tier 1 Grant and a Start-up Grant from Nanyang Technological University. We sincerely thank all volunteers who participated as subjects in our data collection for their time, effort, and support.

%% file: sec/7_supp.tex
\clearpage
\setcounter{page}{1}
\maketitlesupplementary

\noindent\paragraph{Overview.}
This supplementary material is organized as follows. (i) We first describe the overall dataset structure and a series of data compression and acceleration techniques for efficient use of our large-scale \sysname (cf. Sec.~2). (ii) We then present the action design in our dataset and its intended use cases (cf. Sec.~3). (iii) We provide full implementation details for RT-Mesh, all competing methods, the multi-modal fusion setup, and the HAR benchmarks (cf. Sec.~4). (iv) Finally, we offer a more comprehensive assessment of radar-based HMR across different actions and sensing ranges, together with additional visualizations comparing modalities, results under the more challenging S2 and S3 splits (cf. Sec.~5). We also include a demo.mp4 for video visualization.

\section{Ethics Statement}
The \sysname data has been de-identified by a facial mask. The subject recruitment is voluntary, and the involved subject has been informed that the de-identified data will be made publicly available for research purposes. As far as we know, this research does not endanger any person directly. Nevertheless, it is acknowledged that pose estimation and activity recognition research can potentially be used with malicious intent, such as user behavior monitoring.

\section{Dataset Structure}

\begin{figure}[!ht]
\centering
\includegraphics[width=1.\linewidth]{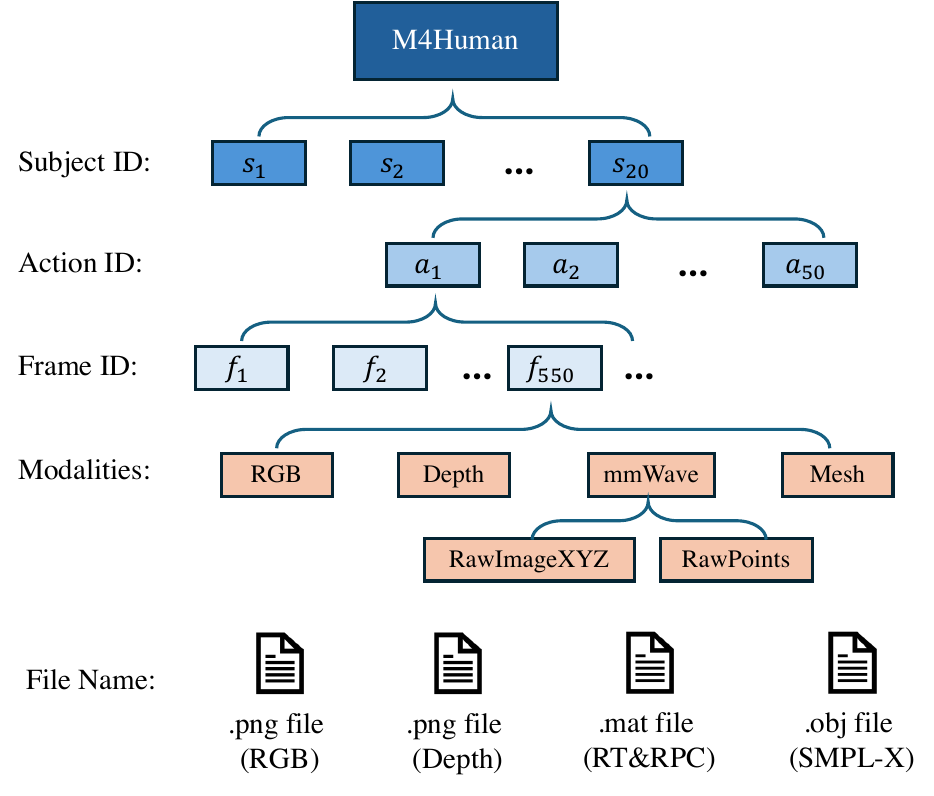}
\caption{The expanded directory of raw \sysname dataset.}
\vspace{-1em}
\label{filestructure}
\end{figure}

\begin{figure*}[!ht]
\centering
\includegraphics[width=1.\linewidth]{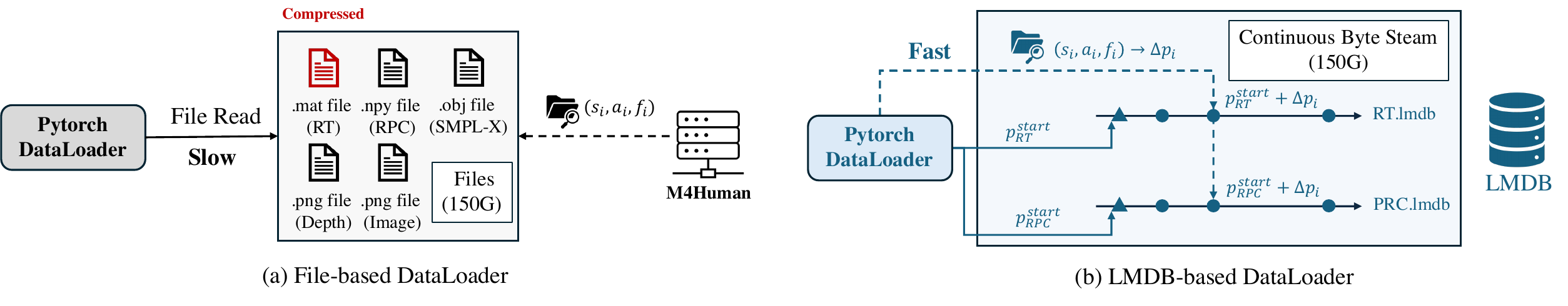}
\caption{(a) A conventional file-system dataloader repeatedly performs file-name lookup and disk I/O on large \texttt{.mat} files (e.g., RT), which quickly becomes a bottleneck at scale.
(b) Our LMDB-based system converts all data into a single contiguous byte stream stored in a memory-mapped database, indexed by the key $(s_i, a_i, f_i)$. At query time, the dataloader first retrieves a base pointer $p^{\text{start}}$ and an offset $\Delta p_i$ associated with the key, and then directly reads the required bytes from disk at $p^{\text{start}} + \Delta p_i$, avoiding expensive directory traversal and repeated file open/close operations.} 

\vspace{-1em}
\label{lmdb}
\end{figure*}

\noindent\paragraph{Overall File Structure of Raw Data.}
We provide two ways to download \sysname: (1) a full archive of the raw data and (2) modality-specific preprocessed archives. 
The full archive preserves the original directory hierarchy so that users can easily inspect the recovered dataset and cross-check samples with the provided annotations (c.f. Fig.~\ref{filestructure}). 
For efficient training, we additionally release preprocessed packages in \texttt{.lmdb} format that is directly compatible with our dataloaders and avoids repeated on-the-fly preprocessing. (c.f. Fig.~\ref{lmdb})

\noindent\paragraph{LMDB-based Data-Loading Acceleration.}
Raw radar tensors are highly I/O-intensive and can severely slow down dataset usage. 
To improve data-loading efficiency and overall usability, we design RT-LMDB, an RT-aware data-loading system built on the Lightning Memory-Mapped Database (LMDB) architecture~\cite{chu2011mdb,henry2019howard}. 
LMDB is an embedded transactional key–value store that keeps arbitrary key–value pairs as byte arrays in a contiguous memory-mapped region on disk, enabling fast random access. 
Similar LMDB-based layouts have been widely adopted in large-scale datasets (e.g., ImageNet~\cite{deng2009imagenet,russakovsky2015imagenet}) to accelerate data loading.

Our pipeline first applies lossless sparsity-aware compression to each RT. 
Instead of storing the full dense voxel grid, we keep only non-zero entries and their voxel indices $(id_x, id_y, id_z, \text{intensity})$, so that the original tensor can be exactly reconstructed by scattering intensities back to the corresponding voxels. 
This compression reduces the storage requirement from approximately $20$\,TB to about $150$\,GB.

However, directly reading compressed files from the file system remains time-consuming, while caching all RTs in RAM is prohibitively expensive and does not scale to multi-GPU training (e.g., 4 GPUs would roughly require $4\times$ more RAM). 
Therefore, we convert all RT files into a single LMDB database, \texttt{RT.lmdb}, where each entry stores the compressed bytes of one frame. 
Each additional modality is stored in its own LMDB (e.g., \texttt{RPC.lmdb}, \texttt{mesh.lmdb}). 
In \sysname, every sample is uniquely identified by the triplet $(s_i, a_i, f_i)$, denoting subject ID, action ID, and frame ID, which we use as the LMDB key. 
As illustrated in Fig.~\ref{lmdb}, the dataloader first queries LMDB to obtain the start pointer $p^{\text{start}}$ of the corresponding byte segment, and then reads the bytes directly from disk. 
Empirically, RT-LMDB achieves random-access speed comparable to keeping all data in RAM, while only storing lightweight pointers in memory, thereby substantially accelerating RT loading for large-scale training. 
Under a single-GPU, single-process setting, the total data-loading time for \sysname{} is reduced from $>18$\,hrs/epoch to $<1$\,hr/epoch.

\noindent\paragraph{Modality Dimensionality.} For storage and I/O efficiency, RGB images are stored with shape $3 \times 640 \times 480$ and depth images with shape $1 \times 640 \times 480$. Radar point clouds (RPC) are stored as tensors of shape $N \times 4$, where $N$ is the number of detected points (typically ranging from 400–600) and each point is represented by $(x,y,z, intensity)$ coordinate and energy response. The high-resolution raw radar tensor (RT) has shape $121 \times 111 \times 31$, corresponding to the spatial resolutions along the $x$, $y$, and $z$ axes, respectively. The value within each voxel corresponds to the intensity.

\section{Detailed Action Set}
As summarized in Table~\ref{tab:actions}, we define 50 actions in three groups: 30 \emph{in-place} daily/rehabilitation exercises, 5 \emph{sit-in-place} chair-based exercises, and 15 \emph{non-in-place} dynamic activities. Across all protocols, the subject–sensor distance ranges from 0.5,m to 4,m. Actions are performed with different facing directions, and subjects are allowed to move within an area of roughly $6,\text{m}^2$. This setup provides diverse motion patterns, enabling a robust evaluation of model performance.

\sysname covers common daily activities and a complete exercise routine, including warm-up, strength training, and post-exercise stretching at different intensity levels. (i) In-place actions. These actions are performed while the subject remains approximately at a fixed location. They are designed to span the main stages of an exercise session. Gentle chest expansions (horizontal/vertical) and left/right trunk twists serve as warm-up and mobility exercises. Squats, various front and side lunges, and jumping or high-knee runs target lower-limb strength and balance. Upper/lower limb extension movements are used as stretching exercises after higher-intensity actions. (ii) Sit-in-place and rehabilitation-oriented actions. Several rehabilitation-oriented patterns are included, such as sideways walking to train lateral stability. Sit-in-place actions are executed on a chair and focus on lower-limb extensions and unilateral arm curls with light loads (e.g., an empty water bottle), mimicking common geriatric and post-surgery rehabilitation routines. (iii) Non-in-place actions. These actions require more complex whole-body coordination and involve substantial translation and direction changes. Examples include straight and curved walking at different speeds, lunge and front-kick walking, as well as sports-like activities such as three-step layups in basketball, badminton with lateral footwork and racket swings, table tennis and volleyball strokes, and free-pace boxing with bouncing footwork and rapid punching.

We expect that this carefully designed action set can support a wide range of real-world applications, including (but not limited to) the following:

\begin{table*}[t]
\centering
\caption{Action set in M4Human.}
\renewcommand{\arraystretch}{1}
\setlength\tabcolsep{5pt}
\resizebox{.9\linewidth}{!}{

    \begin{tabular}{lllllll}
    \toprule
    \textbf{Action ID} & \textbf{Action Name} & \textbf{Protocol} &       & \textbf{Action ID} & \textbf{Action Name} & \textbf{Protocol} \\
    \midrule
    1     & Chest expanding horizontally & In-Place Daily &       & 26    & Right limbs extension & In-Place Rehab \\
    2     & Chest expanding vertically & In-Place Daily &       & 27    &Jumping up & In-Place Daily \\
    3     & Left side twist & In-Place Daily &       & 28    & Tai Chi & In-Place Rehab \\
    4     & Right side twist & In-Place Daily &       & 29    & High knees & In-Place Rehab \\
    5     & Raising left arm & In-Place Daily &       & 30    & Neck rotations while standing with both legs & In-Place Rehab \\
    6     & Raising right arm & In-Place Daily &       & 31    & Slowly stand up and sit down from  a chair & Sit-In-Place Daily \\
    7     & Waving left arm & In-Place Daily &       & 32    & Sit and left leg kick/extension & Sit-In-Place Rehab \\
    8     & Waving right arm & In-Place Daily &       & 33    & Sit and right leg kick/extension & Sit-In-Place Rehab \\
    9     & Picking up things & In-Place Daily &       & 34    & Sit and raise left dumbbell (arm curls) & Sit-In-Place Rehab \\
    10    & Throwing toward left side  & In-Place Daily &       & 35    &Sit and raise right dumbbell (arm curls) & Sit-In-Place Rehab \\
    11    & Throwing toward right side & In-Place Daily &       & 36    & Walk in straight line (fast) & Non-In-Place Daily \\
    12    & Kicking toward left direction using right leg & In-Place Daily &       & 37    & Walk in straight line (slow) & Non-In-Place Daily \\
    13    & Kicking toward right direction using left leg & In-Place Daily &       & 38    & Walk in curves (fast) & Non-In-Place Daily \\
    14    & Bowing forward & In-Place Daily &       & 39    & Walk in curves (slow) & Non-In-Place Daily \\
    15    & Stretching and relaxing in free form & In-Place Daily &       & 40    & Non-in-place Lunge & Non-In-Place Sports \\
    16    & Mark time & In-Place Rehab &       & 41    & Front kick walk & Non-In-Place Sports \\
    17    & Left upper limb extension & In-Place Rehab &       & 42    & Sideways walking & Non-In-Place Daily \\
    18    & Right upper limb extension & In-Place Rehab &       & 43    & Badminton & Non-In-Place Sports \\
    19    & Left front lunge & In-Place Rehab &       & 44    & Table tennis (ping pong) & Non-In-Place Sports \\
    20    & Right front lunge & In-Place Rehab &       & 45    &  Baseball & Non-In-Place Sports \\
    21    & Both upper limbs extension & In-Place Rehab &       & 46    & Volleyball & Non-In-Place Sports \\
    22    & Squat & In-Place Rehab &       & 47    & Free-place Boxing & Non-In-Place Sports \\
    23    & Left side lunge & In-Place Rehab &       & 48    & Walk in straight line \& pick up item & Non-In-Place Daily \\
    24    & Right side lunge & In-Place Rehab &       & 49    & Walk in curve \& pick up item & Non-In-Place Daily \\
    25    & Left limbs extension & In-Place Rehab &       & 50    & Basketball & Non-In-Place Sports \\
    \bottomrule
    \end{tabular}%

    }
    
\label{tab:actions}
\end{table*}

\noindent\paragraph{Privacy-preserving healthcare applications.}
Privacy-preserving sensing is critical for elderly care and disease analysis, as monitored individuals are often unwilling to expose their daily activities, whereas there are strong ethical constraints on releasing patient videos captured by cameras for research purposes. For example, current motion analysis for Parkinson’s disease commonly relies on RGB-D–based skeleton extraction as a compromise for privacy, but such modalities only provide sparse keypoints information~\cite{morgan2023multimodal}. Although depth cameras are more privacy-friendly than RGB by removing facial textures, they still reveal body shape, posture, and facial outlines. They are also affected by clothing bulk (e.g., whether a person is naked) and can potentially be used for re-rendering and re-identification. By contrast, RF sensors, such as mmWave radar, offer truly privacy-preserving sensing. They do not expose the above visual cues, yet still provide abundant motion information, making them a promising modality for data collection and human-motion analysis in healthcare scenarios. Our dataset is the first to demonstrate the potential of extracting 3D dense human meshes for more dynamic activities in an open room, thereby pushing RF sensing toward higher-fidelity perception, better generalizability, and more advanced algorithm design.

\noindent\paragraph{Smart Home and Human Computer Interaction.}
The proposed dataset includes a rich action set covering both in-place and non–in-place daily movements, such as in-place waving, hand raising, throwing, kicking, and non-in-place walking, at different paces/speeds and sitting-standing. These actions naturally support standard action recognition and whole-body gestures, enabling RF-based smart-home applications such as device on/off control, adaptive energy management, and presence/behavior-aware automation. Moreover, the action/gesture samples provide a basis for privacy-preserving HCI, where users can interact with TVs, AR/VR displays, or service robots through RF-sensed body movements without exposing their visual appearance. 

\noindent\paragraph{VR Rendering and Fitness Gaming.}
Thanks to the high-precision marker-based motion capture system and synchronized RGB-D streams, our dataset provides rich annotations for a wide range of 3D rendering and reconstruction tasks. The annotations include 3D SMPL-X human meshes, 3D dense poses, and RGB-textured human meshes, which together enable learning-based 3D human reconstruction from RGB-D, from privacy-preserving mmWave radar, and from multi-modal fusion. Beyond static reconstruction, the diverse action set (covering in-place and non–in-place daily activities, fitness/rehab exercises, and free-space sports-like motions) facilitates the learning of human motion priors, such as coordinated upper–lower limb movements. Consequently, the dataset can support downstream tasks including motion generation, stochastic motion prediction, avatar driving from RF signals, and controllable motion reenactment. This is particularly valuable for VR/AR content creation and fitness gaming, where plausible, temporally coherent human motion is required, as well as for privacy-preserving VR interaction without exposing the user’s visual appearance.

\begin{table*}[t]
\centering
\caption{Radar-based HMR results on \sysname using state-of-the-art indoor human sensing HMR/HPE models. All methods are evaluated on the S1, S2, and S3 splits under four protocols: P1 In-Place (IP), P2 Sit-In-Place (SIP), P3 Non-In-Place (NIP), and ALL (all actions).}
\renewcommand{\arraystretch}{1}
\setlength\tabcolsep{4pt}
\resizebox{1.\linewidth}{!}{


    \begin{tabular}{c|c|c|cccc|cccc|cccc|rl}
    \toprule
    \multirow{2}[2]{*}{\textbf{Modality}} & \multirow{2}[2]{*}{\textbf{Method}} & \multirow{2}[2]{*}{\textbf{Protocol}} & \multicolumn{4}{c|}{\textbf{S1 (Random)}} & \multicolumn{4}{c|}{\textbf{S2 (Cross-Sub)}} & \multicolumn{4}{c|}{\textbf{S3 (Cross-Act)}} & \multicolumn{2}{c}{\multirow{2}[2]{*}{\textbf{Efficiency}}} \\
          &       &       & \textbf{MVE} & \textbf{MJE} & \textbf{MRE} & \textbf{TE} & \textbf{MVE} & \textbf{MJE} & \textbf{MRE} & \textbf{TE} & \textbf{MVE} & \textbf{MJE} & \textbf{MRE} & \textbf{TE} & \multicolumn{2}{c}{} \\
    \midrule
    \midrule
    \multirow{8}[4]{*}{RPC} & \multirow{4}[2]{*}{mmMesh~\cite{xue2021mmmesh}} & \textbf{IP} & 105.9 & 88.0  & 11.0  & 57.2  & 149.4 & 132.6 & 16.1  & 96.2  & 146.0 & 122.9 & 16.1  & 78.5  & Latency (ms): & 3.53 \\
          &       & \textbf{SIP} & 183.3 & 164.2 & 10.7  & 92.8  & 202.8 & 173.5 & 12.2  & 100.5 & 194.5 & 170.9 & 10.8  & 100.3 & Param. (M): & 41.45 \\
          &       & \textbf{NIP} & 201.4 & 170.6 & 15.4  & 105.4 & 226.6 & 188.6 & 17.6  & 113.0 & 223.3 & 187.4 & 18.7  & 115.8 & GFLOPs: & 2.87 \\
          &       & \textbf{ALL} & 132.7 & 112.1 & 11.8  & 70.4  & 170.1 & 147.9 & 16.0  & 100.0 & 173.8 & 146.9 & 16.4  & 91.8  &       &  \\
\cmidrule{2-17}          & \multirow{4}[2]{*}{P4Transformer~\cite{fan2021point}} & \textbf{IP} & 71.5  & 59.0  & 8.0   & 37.3  & 129.3 & 114.3 & 14.9  & 81.1  & 132.3 & 111.3 & 15.9  & 69.9  & Latency (ms): & 7.17 \\
          &       & \textbf{SIP} & 115.2 & 109.3 & 9.5   & 69.1  & 142.8 & 134.6 & 11.9  & 90.0  & 132.0 & 124.0 & 10.3  & 76.9  & Param. (M): & 129.01 \\
          &       & \textbf{NIP} & 139.7 & 121.7 & 13.5  & 77.9  & 180.2 & 158.8 & 17.0  & 103.1 & 184.4 & 157.9 & 18.3  & 96.1  & GFLOPs: & 11.76 \\
          &       & \textbf{ALL} & \textbf{89.5} & \textbf{76.6} & 9.2   & 48.6  & 140.8 & 125.2 & 15.0  & 86.4  & 147.8 & 126.4 & 16.1  & 78.4  &       &  \\
    \midrule
    \midrule
    \multirow{12}[6]{*}{RT} & \multirow{4}[2]{*}{RT-Pose~\cite{ho2024rt}} & \textbf{IP} & 80.0  & 66.5  & 8.7   & 43.0  & 135.9 & 119.7 & 15.3  & 88.6  & 133.7 & 112.0 & 15.7  & 71.6  & Latency (ms): & 39.58 \\
          &       & \textbf{SIP} & 130.2 & 126.0 & 9.9   & 81.6  & 152.4 & 145.1 & 12.0  & 100.9 & 139.5 & 132.6 & 10.5  & 86.1  & Param. (M): & 6.05 \\
          &       & \textbf{NIP} & 158.5 & 139.3 & 14.2  & 92.2  & 188.4 & 164.4 & 17.0  & 109.0 & 196.0 & 169.0 & 18.4  & 110.4 & GFLOPs: & 50.67 \\
          &       & \textbf{ALL} & 100.7 & 87.0  & 9.9   & 56.7  & 148.1 & 131.2 & 15.3  & 93.9  & 152.8 & 131.0 & 16.0  & 84.6  &       &  \\
\cmidrule{2-17}          & \multirow{4}[2]{*}{RETR~\cite{yataka2024retr}} & \textbf{IP} & 73.2  & 58.8  & 7.5   & 34.8  & 159.4 & 138.9 & 16.6  & 95.4  & 143.0 & 118.3 & 16.5  & 70.9  & Latency (ms:) & 17.87 \\
          &       & \textbf{SIP} & 133.6 & 126.2 & 10.0  & 80.4  & 169.4 & 156.7 & 12.1  & 105.7 & 153.6 & 143.6 & 10.8  & 93.1  & Param. (M): & 52.42 \\
          &       & \textbf{NIP} & 162.6 & 140.4 & 14.2  & 90.1  & 206.1 & 178.5 & 17.4  & 114.1 & 207.0 & 177.1 & 18.8  & 112.3 & GFLOPs: & 3.01 \\
          &       & \textbf{ALL} & 97.1  & 81.8  & \textbf{9.1} & 50.4  & 169.7 & 148.6 & 16.3  & 100.1 & 163.1 & 138.3 & 16.6  & 85.4  &       &  \\
\cmidrule{2-17}          & \multicolumn{1}{c|}{\multirow{4}[2]{*}{RT-Mesh (Ours)}} & \textbf{IP} & 72.4  & 59.1  & 8.3   & 36.2  & 123.6 & 109.6 & 14.7  & 82.1  & 128.5 & 107.2 & 15.3  & 68.1  & Latency (ms): & 2.74 \\
          &       & \textbf{SIP} & 118.1 & 112.2 & 9.7   & 68.4  & 138.5 & 130.8 & 12.0  & 88.4  & 126.5 & 120.0 & 10.4  & 73.9  & Param. (M): & 63.25 \\
          &       & \textbf{NIP} & 142.0 & 123.1 & 13.8  & 77.1  & 173.6 & 151.8 & 16.9  & 98.8  & 178.2 & 152.6 & 18.0  & 92.8  & GFLOPs: & 2.60 \\
          &       & \textbf{ALL} & 90.9  & 77.2  & 9.6   & \textbf{47.6} & \textbf{135.1} & \textbf{120.2} & \textbf{14.9} & \textbf{86.1} & \textbf{143.1} & \textbf{122.0} & \textbf{15.6} & \textbf{76.0} &       &  \\
    \bottomrule
    \end{tabular}%

    }
    
\label{tab:main}
\end{table*}

\begin{figure*}[!ht]
\centering
\includegraphics[width=.8\linewidth]{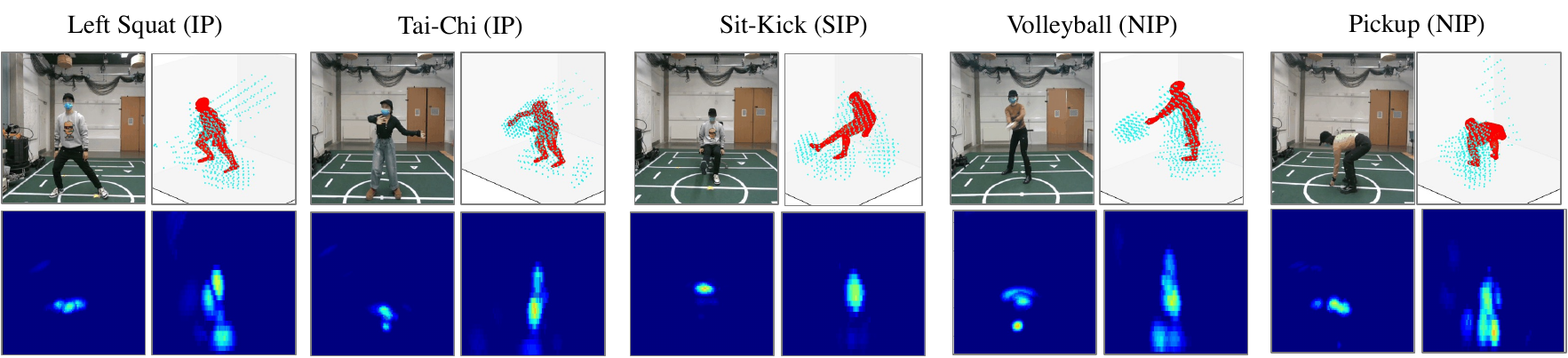}
\caption{Sample visualizations of different modalities and ground-truth annotations. Our calibration system achieves good alignment between RGB-D, radar modalities, and the ground-truth meshes.}
\label{baseline}
\end{figure*}

\section{Benchmark Implement Details}

\subsection{Overall Training Specifications}
Unless specified otherwise, all methods share the same HMR prediction head to regress SMPL-X parameters and follow a unified training setup. All models are implemented in PyTorch and trained for 100 epochs with a batch size of 64. We use the Adam optimizer~\cite{kingma2014adam} with a learning rate of $2\times10^{-4}$, momentum $0.9$, and a learning-rate decay factor of $0.1$ applied every 5 epochs. We further apply norm-based gradient clipping to stabilize training. All methods are optimized with a unified mesh reconstruction loss $\mathcal{L}_{\text{mesh}}$, with loss weights set to $\lambda_\alpha = 1$, $\lambda_\beta = 0.3$, $\lambda_\tau = 10$, $\lambda_\theta = 15$, and $\lambda_g = 0.5$. These hyperparameters are chosen empirically; more advanced tuning strategies (e.g., Bayesian optimization) may further improve performance and are left for future work. All experiments are conducted on a local Ubuntu 20.04 server equipped with 4 NVIDIA RTX 3090 GPUs, an Intel Xeon(R) Platinum 8474C processor (15 cores), and 128\,GB RAM. Training RT-Mesh with 4 GPUs in parallel takes approximately one day.

\subsection{Implementation of RT-Mesh}

\noindent\textbf{BEV 2D Localization.} We reshape $X_{\text{RT}}$ into a BEV tensor along the $X\text{–}Y$ plane by concatenating the remaining axes into channels, i.e., $C_{\text{2D}} = Z \times T$. We adopt BEV since $(X,Y)$ provides the highest spatial resolution for accurate human localization. The BEV tensor is encoded by 2D convolutions with downsampling ratio $N_1$, resulting in a feature map that is patchified into $\frac{X}{N_1}\times \frac{Y}{N_1}$ tokens. A lightweight 2D transformer performs patch-wise self-attention to amplify human responses. The output tokens are globally pooled and flattened into a unified BEV descriptor, from which we regress the human’s BEV coordinates $(\hat{x},\hat{y})$.

\noindent\textbf{Local 3D Regression.} Centered at $(\hat{x},\hat{y})$, we crop a fixed-size 3D RoI $(\Delta X,\Delta Y,\Delta Z){=}(24,24,31)$ from $X_{\text{RT}}$. This compact RoI enables fast processing and reduces clutter/multipath outside the human foreground. The cropped 3D tensor passes through an RoI-specific 3D transformer stack: a short 3D convolutional stem with downsampling ratio $N_2$ captures local 3D structure, the features are patchified, and a 3D transformer applies high-resolution self-attention. The output tokens are globally max-pooled and flattened to form the final 3D mesh feature. Finally, an MLP-based HMR head regresses SMPL-X parameters $(\hat{\alpha},\hat{\beta},\hat{\tau},\hat{\theta},\hat{g})$ for reconstructing the SMPL-X mesh.

\noindent\textbf{Training Objective.} We jointly optimize BEV localization and 3D mesh regression:
\begin{equation}
\mathcal{L} \;=\; \lambda_{2D}\,\mathcal{L}_{2D} \;+\; \lambda_{\text{mesh}}\,\mathcal{L}_{\text{mesh}}~,
\end{equation}
where $\mathcal{L}_{2D} = \lVert\hat{x}-\tau[0]\rVert_2+\lVert\hat{y}-\tau[1]\rVert_2$ supervises 2D BEV localization. The 3D mesh loss is the weighted sum over SMPL-X components:
\begin{equation}
\begin{aligned}
\mathcal{L}_{\text{mesh}} &= \lambda_{\theta}\,\mathcal{L}_{\text{rot}}(\hat{\theta},\theta)
\;+\; \lambda_{\alpha}\,\mathcal{L}_{\text{rot}}(\hat{\alpha},\alpha)
\;+\; \lambda_{\beta}\,\lVert \hat{\beta}-\beta \rVert_2^2 \\
&\quad+\; \lambda_{\tau}\,\lVert \hat{\tau}-\tau \rVert_1
\;+\; \lambda_{g}\,\mathrm{BCE}(\hat{g},g)~,
\end{aligned}
\end{equation}
where $\mathcal{L}_{\text{rot}}$ denotes a geodesic rotation loss~\cite{zhou2019continuity} and $\mathrm{BCE}(\cdot)$ is the binary cross-entropy loss.

\begin{figure*}[!ht]
\centering
\includegraphics[width=1.\linewidth]{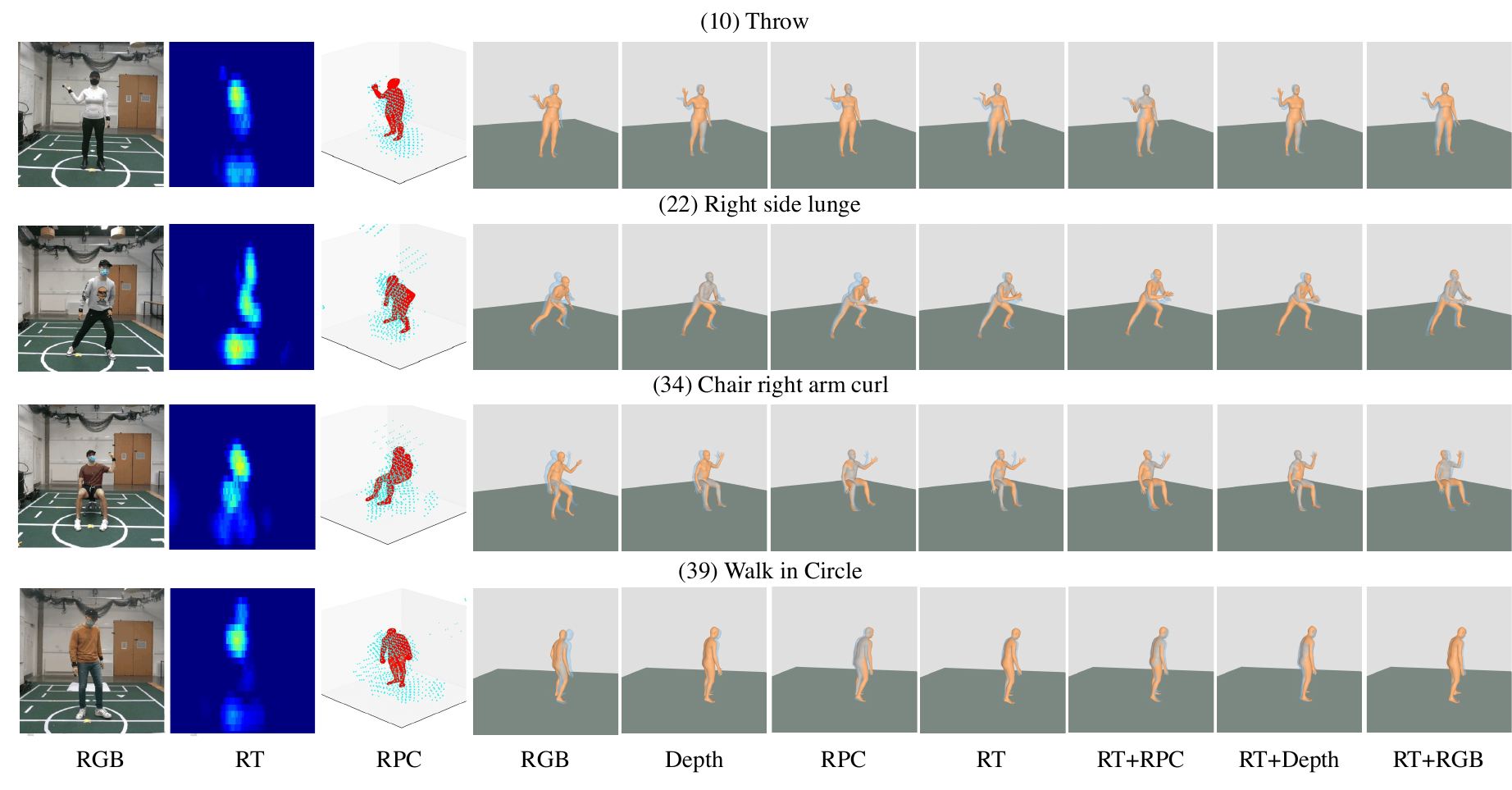}
\caption{Visualization of single-modality predictions and multi-modal fusion. Predicted meshes are shown in orange and ground truth in blue; higher overlap indicates higher accuracy. The performance gap between line-of-sight (LoS) RGB-D and radio-frequency (RF) modalities is smaller than expected, thanks to the high-resolution radar in \sysname, making radar-based HMR feasible. Notably, RGB-only HMR struggles with accurate depth estimation, leading to larger reconstruction errors.}
\label{modality1}
\end{figure*}

\subsection{Implementations of Single-Modal Methods}

\noindent\paragraph{RGB.}
We use Detectron2~\cite{wu2019detectron2} to detect human-centered square bounding boxes, resized to $224\times224$. The cropped RGB patches are passed to a pretrained TokenHMR~\cite{dwivedi2024tokenhmr} model to predict SMPL-X parameters $(\hat{\alpha}_{\mathrm{img}},\hat{\beta}_{\mathrm{img}},\hat{\theta}_{\mathrm{img}},\hat{g}_{\mathrm{img}})$, representing global orientation, body shape, pose, and gender, respectively, under the cropped-image coordinate system.

To localize the subject under the full-image coordinate, the model additionally predicts a normalized triplet $(\hat{s},\hat{u},\hat{v})$, where $\hat{s}$ is the weak-perspective scale (monotonically related to inverse depth) and $(\hat{u},\hat{v})\in[-1,1]$ are normalized crop coordinates. To undo the crop to obtain full-image pixels, we leverage (i) the human detector's outputs: box size $b$ and box center $(c_x,c_y)$, and (ii) the camera intrinsics: focal length $f$ (pixels) and principal point $(W/2,H/2)$. The human center pixels is computed as: 
\begin{equation}
x_{\mathrm{pix}} = c_x + \frac{b_s}{2}\,\hat{u}, \
y_{\mathrm{pix}} = c_y + \frac{b_s}{2}\,\hat{v}, \
\hat{z}_{\mathrm{img}}=\frac{2f}{b_s},
\end{equation}
where $b_s = b *\hat{s}$ is the effective box size and $\hat{z}_{\mathrm{img}}$ is the estimated depth. We then apply pinhole back-projection with the weak-to-perspective depth relation to obtain full translation $\hat{\tau}_{img}=(\hat{x}_{\mathrm{img}},\hat{y}_{\mathrm{img}},\hat{z}_{\mathrm{img}})$ under the world coordinates:
\begin{equation}
\hat{x}_{\mathrm{img}}=\frac{x_{\mathrm{pix}}-\tfrac{W}{2}}{f}\,\hat{z}_{\mathrm{img}},\qquad
\hat{y}_{\mathrm{img}}=\frac{y_{\mathrm{pix}}-\tfrac{H}{2}}{f}\,\hat{z}_{\mathrm{img}}.
\end{equation}

For the best performance, we fine-tune the prediction heads for $(\hat{\alpha}_{\mathrm{img}},\hat{\beta}_{\mathrm{img}},\hat{\tau}_{\mathrm{img}},\hat{g}_{\mathrm{img}})$ on our dataset to improve depth and localization accuracy. We keep the VQ-based pose tokenizer/decoder fixed, as it is pretrained on the large-scale AMASS dataset and provides strong prior performance. This fine-tuning improves the MVE from 190 mm to 97 mm on our benchmark, as reported in the main paper.

\begin{figure*}[!ht]
\centering
\includegraphics[width=.8\linewidth]{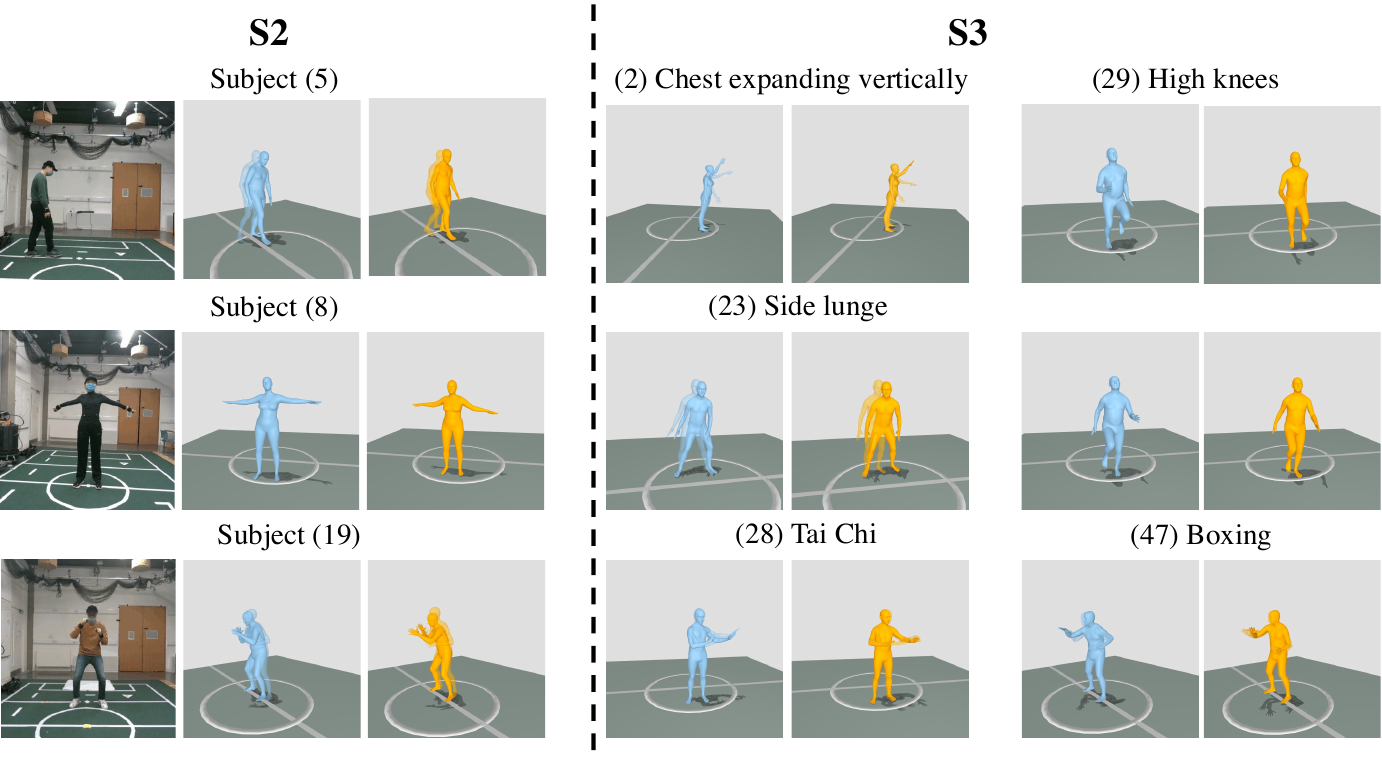}
\caption{Visualization of RT-Mesh under S2 (cross-subject) and S3 (crpss-action) split. RT-Mesh demonstrates good generalization to unseen subjects and actions.}
\label{s2s3}
\end{figure*}

\begin{figure}[!ht]
\centering
\includegraphics[width=\linewidth]{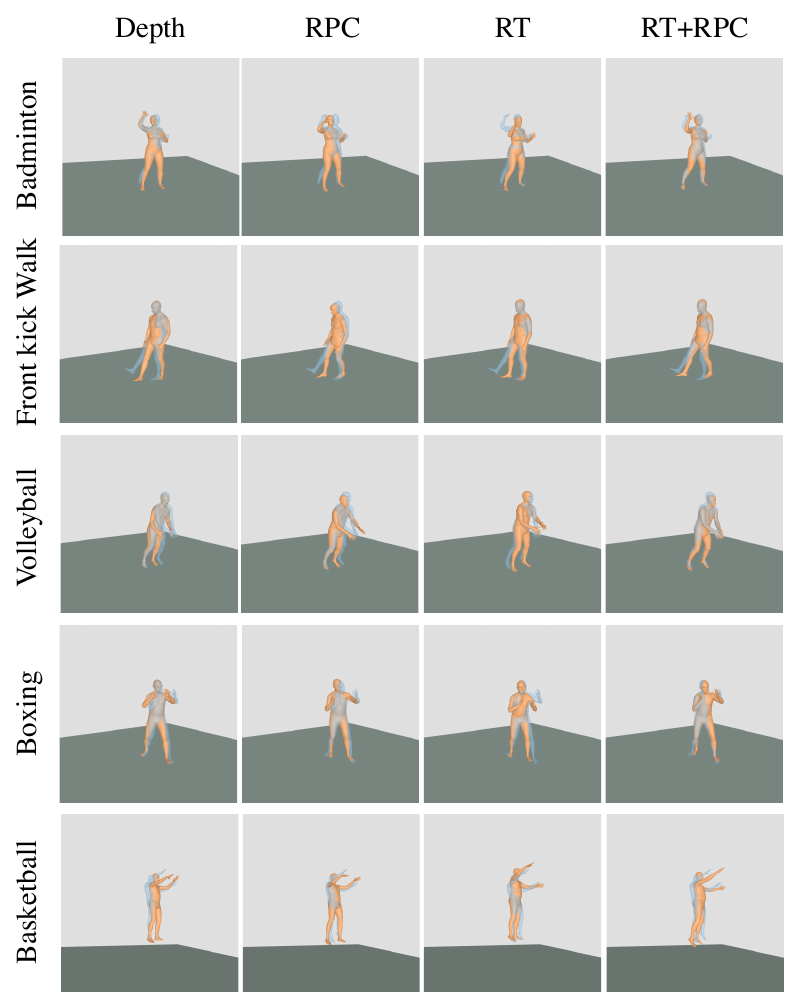}
\caption{Visualization of RT–RPC fusion versus depth-based HMR under dynamic non-in-place actions. Predicted meshes are shown in orange and ground truth in blue; a higher overlap indicates higher accuracy. Fusing RT and RPC consistently improves radar-based HMR, narrowing the performance gap to the depth modality.}
\label{modality2}
\end{figure}

\noindent\paragraph{Depth.}
Following the same pinhole back-projection used in the RGB pipeline, depth frames (which provide measured rather than estimated depth) are back-projected with camera intrinsics $K$ to obtain per-pixel dense 3D points in the camera/world frame, yielding a dense point cloud. We then adopt a widely adopted point-based encoder, P4Transformer~\cite{fan2021point}, to process the depth-derived point cloud, which is adopted in prior multimodal benchmark mmBody~\cite{chen2022mmbody} and LiDAR-based HMR benchmark (e.g., RELI11D). For a fair comparison with the RPC modality, we use the same HMR prediction head to regress SMPL-X parameters and optimize with the same mesh loss $\mathcal{L}_{\text{mesh}}$.

\noindent\paragraph{RPC and RT.}
For RPC preprocessing, we follow prior designs~\cite{yang2023mm,an2022fast,rahman2024mmvr} and adopt a sliding window that aggregates $T{=}4$ adjacent frames into a single input, which helps alleviate point-cloud sparsity and occasional body-part miss-detections in individual frames. Since the number of points varies across frames, we apply zero-padding to a fixed size of 1{,}000 points per aggregated frame, yielding a unified tensor shape for efficient PyTorch batching. We use P4Transformer~\cite{fan2021point} as the point-cloud baseline, as it achieves state-of-the-art performance on previous RPC-based HMR benchmarks (e.g., mmBody~\cite{chen2022mmbody}) and is widely employed in point-based HMR (e.g., LiDAR). For RT preprocessing, we adopt the same $T{=}4$ sliding-window strategy to ensure a fair comparison, resulting in a 4D radar tensor. For RETR, we reshape the 4D tensor $X_{\mathrm{RT}}$ into two views, $X \times Y \times (Z \times T)$ for the horizontal (BEV) view and $Y \times Z \times (X \times T)$ for the vertical ($XZ$) view, and follow the official transformer encoder–decoder design for multi-view feature fusion. For RT-Pose, which relies on 3D convolutions and an HRNet-style head, we treat the temporal dimension $T$ as the input feature (channel) dimension.

\subsection{Implementation of Multi-modal Fusion}
We adopt a simple feature-level fusion strategy to clearly assess the benefit of combining modalities. Let $f^{(m)} \in \mathbb{R}^{d_m}$ denote the final encoder-extracted global feature from modality $m$ (e.g., RGB, depth, RPC, or RT), where $d_m = 1024$ is the feature dimension. Before being fed into the HMR prediction head, we concatenate the features from two modalities along the channel dimension:
\[
f_{\text{fuse}} = \bigl[f^{(m_1)} \,\|\, f^{(m_2)}\bigr] \in \mathbb{R}^{2d_m}.
\]
The fused feature $f_{\text{fuse}}$ is then input to the same HMR prediction head architecture as in the single-modality case, with only the input dimension changed to $2d_m$ to match the fused feature size.

For fairness, we keep all modality backbones, the mesh loss $\mathcal{L}_{\text{mesh}}$, and the overall training setup identical to the unimodal experiments. This minimal design ensures that the performance gains arise from the inclusion of additional modalities rather than from altered training or loss configurations. We expect and encourage that more advanced multi-modal fusion mechanisms (e.g., attention-based fusion, gating, or cross-modal transformers) could further improve performance and leave such designs for future research.

\begin{figure}[!b]
\centering
\includegraphics[width=\linewidth]{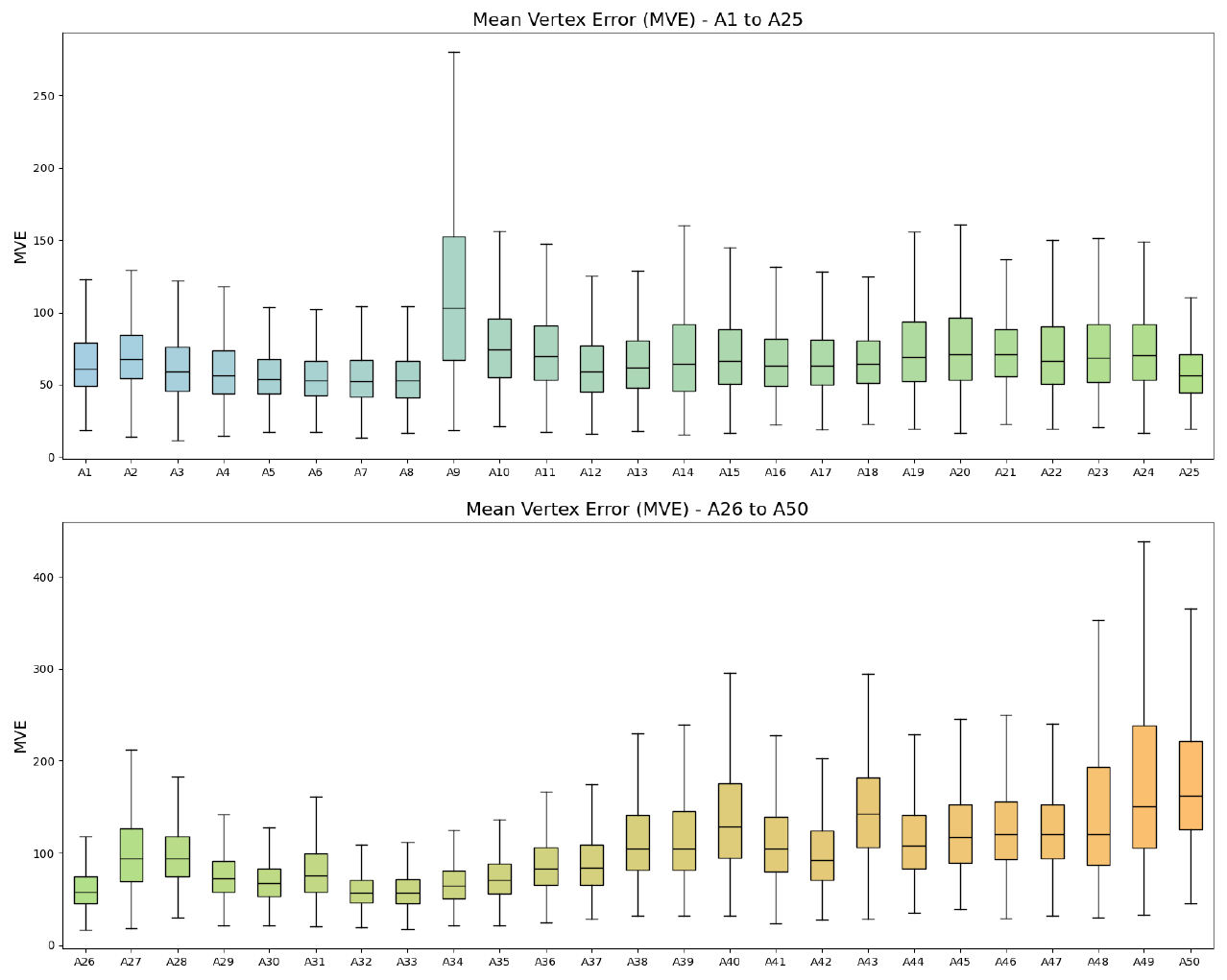}
\caption{Mean Vertex Error (MVE) across different action types. Dynamic non-in-place actions exhibit higher MVE, highlighting their increased difficulty and suggesting the need for more advanced motion modeling and stronger prior knowledge.}
\label{actions_error}
\end{figure}

\subsection{Implementation of HAR}
\sysname can also be used for skeleton-based human action recognition (HAR) across different modalities. Here, we construct a new HAR benchmark based on 3D skeletons extracted from our predicted SMPL-X meshes. Following the official splits used for HMR, we evaluate HAR under S1 (random split) and S2 (cross-subject split). For each sequence in S1 and S2, we first run HMR and then extract 3D joint trajectories from the predicted meshes. Each sequence is segmented into clips of 64 consecutive frames (approximately 5 seconds), yielding around \textit{14,000} training clips, \textit{1,000} validation clips, and \textit{3,200} testing clips. 

\begin{figure}[!ht]
\centering
\includegraphics[width=1.\linewidth]{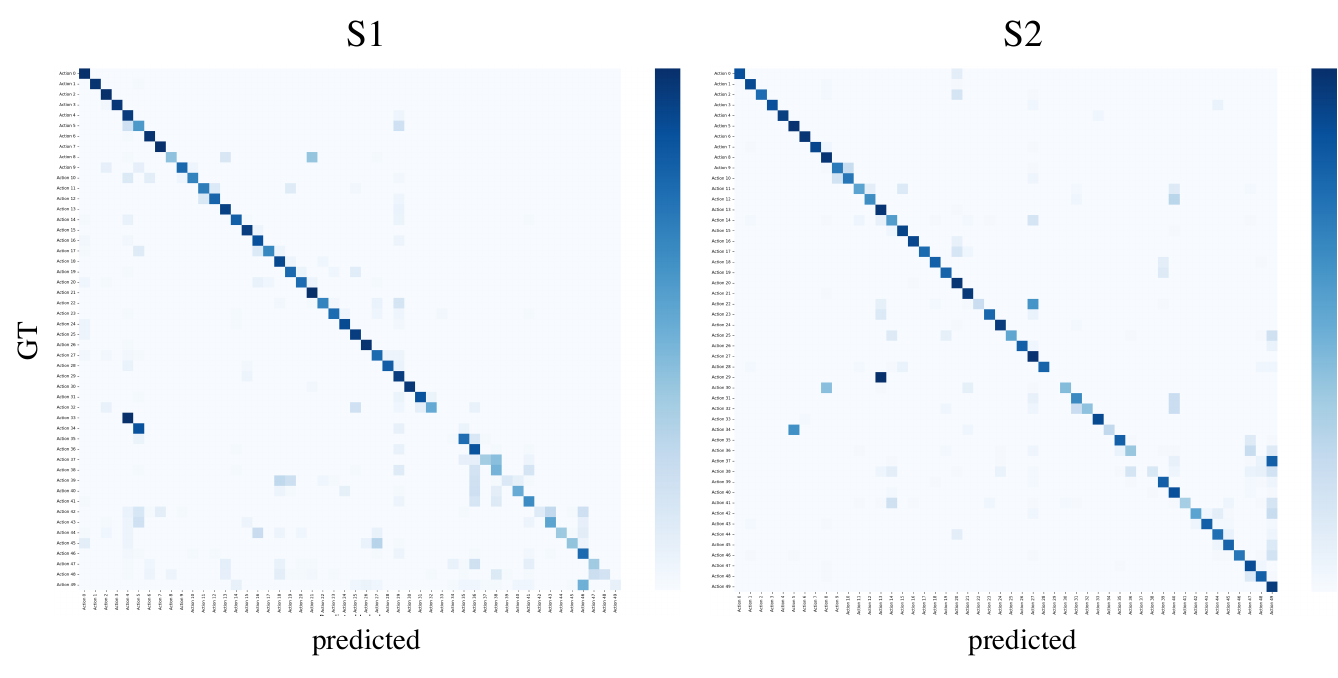}
\caption{Confusion Matrix visualization for RT-based HAR using the best performing AGCN, under S1 and S2 splits.}
\vspace{-1em}
\label{confusion}
\end{figure}

We implement several representative skeleton-based HAR methods: a simple 2D CNN~\cite{du2015skeleton} operating on the temporal–joint (T$\times$J) dimension, AGCN~\cite{shi2019two} (CVPR'19), and BlockGCN~\cite{zhou2024blockgcn} (CVPR'24). For AGCN and BlockGCN, we initialize from models pretrained on NTU-RGBD and fine-tune them on our dataset. We train the model for $1{,}000$ epochs with a batch size of $32$, using the Adam~\cite{kingma2014adam} optimizer with momentum $0.9$, an initial learning rate of $0.01$, and weight decay of $10^{-4}$. We report Top-1 and Top-5 classification accuracy as our evaluation metrics. As shown in Fig.~\ref{confusion}, the confusion matrices obtained using skeletons derived solely from the RT modality exhibit strong classification performance on both S1 and S2, indicating that our radar-predicted meshes are accurate enough to support challenging downstream HAR tasks.

\section{More Qualitative Visualization}

\subsection{Qualitative Comparison Across Modalities}
As illustrated in Fig.~\ref{modality1}, we visualize HMR results for different single modalities (RGB, depth, RPC, RT) as well as their fused predictions. Predicted meshes are overlaid in orange and ground truth in blue. Due to the lack of explicit depth measurements, RGB-only HMR often exhibits noticeable depth offsets, which leads to biased global localization and reduced overlap with the ground-truth meshes. In contrast, depth-based HMR yields more accurate reconstructions, benefiting from high-precision per-pixel depth. RPC- and RT-based methods achieve comparable localization quality, even for challenging actions such as side lunges, demonstrating that high-resolution mmWave radar can support fine-grained body pose recovery. Moreover, multi-modal fusion generally outperforms single-modality models: in particular, RT+RPC fusion produces visibly more accurate meshes than either RT or RPC alone. Additional examples of dynamic non-in-place motions are shown in Fig.~\ref{modality2}, where RT+RPC consistently surpasses single-modality radar and approaches the accuracy of depth-based HMR. These qualitative results indicate that radar is capable of high-fidelity human sensing. At the same time, as also seen in Fig.~\ref{modality1}, radar representations do not reveal identifiable appearance or background details, making RF-based HMR a promising solution for privacy-sensitive applications.

\subsection{HMR Visualization for Cross-Subject and Cross-Action}
As illustrated in Fig.~\ref{s2s3}, we present qualitative results for the more challenging S2 (cross-subject) and S3 (cross-action) settings. In both cases, we observe that body motion and pose dynamics can be reasonably well recovered, whereas the estimated body shape parameters $\boldsymbol{\beta}$ are less accurate and sometimes inconsistent across frames. We hypothesize that this limitation stems from the lack of rich appearance cues and fine-grained geometric correspondences in radar data compared to RGB-D, which also contributes to its privacy-preserving nature.

For S3, our model shows promising generalization to unseen daily activities such as vertical chest expansion and high knees, even though these actions do not appear in the training set. This suggests that the proposed RT-Mesh is able to capture meaningful spatio-temporal body dynamics that transfer to novel motion patterns, making it suitable for generalizable human monitoring. However, the model still struggles with more complex motions, such as side lunges or highly dynamic non-in-place boxing, which require a deeper understanding of human kinematics and body coordination. We believe that the future integration of stronger human priors and more expressive motion models will improve generalization to such challenging actions.

\begin{figure}[!ht]
\centering
\includegraphics[width=1.\linewidth]{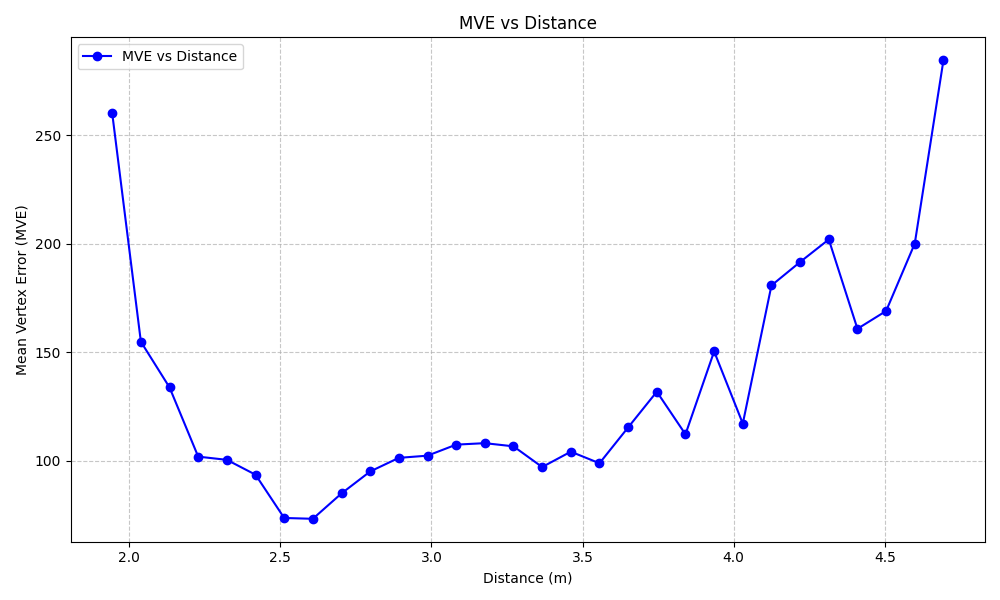}
\caption{Mean Vertex Error (mm) versus sensing distance (m). The optimal sensing range is around 2.0 to 4.0 m. }
\vspace{-1em}
\label{mve_vs_distance}
\end{figure}

\subsection{HMR Performance Across Actions and Sensing Distances}
As shown in Fig.~\ref{actions_error}, we report Mean Vertex Error (MVE) for all 50 actions using box plots grouped by action type. Our models perform well on simple in-place and sit-in-place exercises, with errors typically below 100,mm. In contrast, non-in-place dynamic actions are noticeably more challenging, as they involve more complex motion coordination and kinematics and therefore require richer motion understanding and stronger priors, which we leave for future study.

Fig.~\ref{mve_vs_distance} further plots MVE as a function of sensing distance, illustrating how radar-based HMR accuracy varies with subject–sensor range. Performance generally degrades at larger distances, where fewer foreground points are captured, and also at very close ranges, where the human body can exceed the field of view. The best accuracy is obtained at medium distances, approximately 2.0–4.0 m. We expect that more advanced algorithmic designs could further extend the effective sensing range.